\PassOptionsToPackage{table}{xcolor}
\documentclass[]{selfevolagent}

\usepackage{microtype}
\microtypesetup{expansion=false}
\usepackage{graphicx}
\usepackage{subcaption}
\usepackage{booktabs}
\usepackage{hyperref}
\usepackage{amsmath}
\usepackage{amssymb}
\usepackage{mathtools}
\usepackage{amsthm}
\usepackage{latexsym}
\usepackage[T1]{fontenc}
\usepackage[utf8]{inputenc}
\usepackage{multirow}
\usepackage{balance}
\usepackage{pifont}
\usepackage{makecell}
\usepackage{filecontents}
\usepackage{bbm}
\usepackage{pgfplots}
\usetikzlibrary{patterns}
\usepackage[inline,shortlabels]{enumitem}
\usepackage{natbib}
\usepackage{csquotes}
\usepackage{url}
\usepackage{svg}
\usepackage{bm}
\usepackage{graphics}
\usepackage{array}
\usepackage[english]{babel}
\usepackage{textcomp}
\usepackage{siunitx}
\usepackage[normalem]{ulem}
\useunder{\uline}{\ul}{}
\usepackage{titletoc}
\usepackage{tikz}
\usepackage{arydshln}
\usepackage[most]{tcolorbox}
\usepackage{algorithm}
\usepackage{algpseudocode}
\usepackage{fontawesome5}
\usepackage{simpleicons}
\usepackage{wrapfig}
\usepackage{adjustbox}
\usepackage{listings}
\usepackage[textsize=tiny]{todonotes}
\usetikzlibrary{tikzmark}

\makeatletter
\newcommand*\myfontsize{%
\@setfontsize\myfontsize{7}{8}%
}
\makeatother

\graphicspath{{Images/}}

\definecolor{geminiBlue}{HTML}{8E8ED7}
\definecolor{qwenBlue}{HTML}{78A2E0}
\definecolor{myred}{rgb}{0.7, 0.3, 0.0}
\definecolor{myblue}{HTML}{0a41b8}
\definecolor{mygreen}{HTML}{056b34}
\definecolor{mypurple}{HTML}{5d1e8b}
\definecolor{mmskillrow}{RGB}{236,248,241}
\definecolor{dividergray}{RGB}{240,240,240}
\definecolor{ourslavender}{RGB}{239,237,255}
\definecolor{headergray}{RGB}{250,250,250}
\definecolor{codegreen}{rgb}{0,0.6,0}
\definecolor{codegray}{rgb}{0.5,0.5,0.5}
\definecolor{codepurple}{rgb}{0.58,0,0.82}
\definecolor{backcolour}{rgb}{0.95,0.95,0.92}
\definecolor{startBlue}{HTML}{1628a7}
\definecolor{endPurple}{HTML}{8b16aa}
\definecolor{wx}{RGB}{54, 89, 170}

\newcommand{\mmskillhl}{\cellcolor{mmskillrow}}

\newcommand{\huggingfaceicon}{\raisebox{-0.16ex}{\scalebox{0.94}{\simpleicon{huggingface}}}}
\newcommand{\corrauth}{\text{\raisebox{-0.12ex}{\scalebox{0.78}{\faEnvelope}}}}

\lstdefinestyle{pystyle}{
    backgroundcolor=\color{backcolour},
    commentstyle=\color{codegreen},
    keywordstyle=\color{magenta},
    numberstyle=\tiny\color{codegray},
    stringstyle=\color{codepurple},
    basicstyle=\ttfamily\small,
    breakatwhitespace=false,
    breaklines=true,
    captionpos=b,
    keepspaces=true,
    numbers=left,
    numbersep=5pt,
    showspaces=false,
    showstringspaces=false,
    showtabs=false,
    tabsize=4
}
\theoremstyle{plain}

\theoremstyle{definition}

\theoremstyle{remark}

\newtcolorbox[auto counter, number within=section]{promptbox}[2][]{%
  colback=white,
  colframe=myblue,
  width=\textwidth,
  arc=2mm,
  title={\normalsize\faInfoCircle\hspace{0.5em}#2},
  breakable,
  fonttitle=\bfseries\Large,
  fontupper=\small,
  drop shadow southeast,
  top=2mm,
  bottom=2mm,
  before skip=3mm,
  after skip=3mm,
  boxrule=0.5mm,
  #1
}

\title{MMSkills: Towards Multimodal Skills for General Visual Agents}
\author[1,2\ddagger*]{Kangning Zhang}
\author[1,2\ddagger*]{~Shuai Shao}
\author[1,2*]{~Qingyao Li}
\author[1]{~Jianghao Lin}
\author[1]{~Lingyue Fu}
\author[3]{~Shijian Wang}
\author[2,\corrauth]{\\Wenxiang Jiao}
\author[2,\corrauth]{~Yuan Lu}
\author[1,\corrauth]{~Weiwen Liu}
\author[1,\corrauth]{~Weinan Zhang}
\author[1,\corrauth]{~Yong Yu}

\affiliation[1]{Shanghai Jiao Tong University}
\affiliation[2]{Xiaohongshu Inc.}
\affiliation[3]{Southeast University}

\contribution[*]{Work done during internship at Xiaohongshu Inc.}
\contribution[\ddagger]{Equal contribution}
\contribution[\corrauth]{Corresponding authors}

\metadata[\raisebox{-0.18ex}{\scalebox{0.89}{\faEnvelope}}~Contact]{zhangkangning@sjtu.edu.cn, wenxiangjiaonju@gmail.com, liuww@sjtu.edu.cn}
\metadata[{\faGithub}~Code \& Demo]{\href{https://github.com/zkangning/MMSkills_for_Visual_Agents}{\nolinkurl{https://github.com/zkangning/MMSkills_for_Visual_Agents}}}
\metadata[{\huggingfaceicon}~Skill Source]{\href{https://huggingface.co/datasets/zhangkangning/mmskills}{\nolinkurl{huggingface.co/datasets/zhangkangning/mmskills}}}
\metadata[{\faGlobe}~Website]{\href{https://zkangning.github.io/MMSkills_for_Visual_Agents/}{\nolinkurl{https://zkangning.github.io/MMSkills_for_Visual_Agents/}}}

\abstract{
Reusable skills have become a core substrate for improving agent capabilities, yet most existing skill packages encode reusable behavior primarily as textual prompts, executable code, or learned routines. For visual agents, however, procedural knowledge is inherently multimodal: reuse depends not only on what operation to perform, but also on recognizing the relevant state, interpreting visual evidence of progress or failure, and deciding what to do next. We formalize this requirement as \emph{multimodal procedural knowledge} and address three practical challenges: (I) \textbf{what} a multimodal skill package should contain; (II) \textbf{where} such packages can be derived from public interaction experience; and (III) \textbf{how} agents can consult multimodal evidence at inference time without excessive image context or over-anchoring to reference screenshots. We introduce \emph{MMSkills}, a framework for representing, generating, and using reusable multimodal procedures for runtime visual decision making. Each MMSkill is a compact, state-conditioned package that couples a textual procedure with runtime state cards and multi-view keyframes. To construct these packages, we develop an agentic trajectory-to-skill Generator that transforms public non-evaluation trajectories into reusable multimodal skills through workflow grouping, procedure induction, visual grounding, and meta-skill-guided auditing. To use them, we introduce a branch-loaded multimodal skill agent: selected state cards and keyframes are inspected in a temporary branch, aligned with the live environment, and distilled into structured guidance for the main agent. Experiments across GUI and game-based visual-agent benchmarks show that MMSkills consistently improve both frontier and smaller multimodal agents, suggesting that external multimodal procedural knowledge complements model-internal priors.
}

\begin{document}

\maketitle


\section{Introduction}
\label{sec:introduction}

Skills have become one of the central abstractions for building useful agents: recent systems store reusable behaviors as prompts, code, execution graphs, or learned routines that can be retrieved and composed later \citep{wang2023voyager,zheng2025skillweaver,chen2026cuaskill,wang2026skillx}. Despite differences in implementation, these skills largely share a common representational assumption: reusable knowledge can be expressed as a textual or code-level specification of actions. This design is effective when the relevant state can be adequately abstracted in language, but it is insufficient for multimodal agents whose decisions depend on visual evidence. For such agents, reusable experience must specify not only what operation to perform, but also how to recognize the relevant state, and how visual evidence should guide the next decision.
A desktop agent may know the correct operation but fail to recognize that a dialog is not yet ready; a game agent may know the intended goal but still require visual cues to distinguish progress from completion.
This observation is consistent with human procedural learning, where visual information can complement verbal explanations \citep{mayer2009multimedia}.
Consequently, text-only skills become verbose yet underspecified, whereas demonstrations preserve visual context but are lengthy, instance-specific, and difficult to adapt.

This gap suggests the need for \emph{multimodal procedural knowledge}: reusable guidance that binds action procedures to the visual evidence and state-dependent decisions required for applying them.
Such knowledge is not simply a text skill with screenshots attached.
To be reusable, it must specify what procedure is being reused, when the procedure should or should not be used, which visible cues matter, and which evidence verifies progress, failure, or completion.
Turning this requirement into practical multimodal skill libraries raises three central challenges:
\begin{itemize}[leftmargin=*]
    \item \textbf{Representation.} What should a multimodal skill package contain, and how should it bind procedures, visible, and verification cues into a coherent reusable unit?
    \item \textbf{Generation.} Where can such packages be derived from, if they must use public non-evaluation interaction experience rather than hand-written examples or raw demonstration replay?
    \item \textbf{Utilization.} How can an agent consult multimodal skill evidence at inference time while avoiding excessive image context, distracting state descriptions, and over-anchoring to reference screenshots?
\end{itemize}

We propose \textbf{MMSkills}, a framework for representing, generating, and utilizing reusable multimodal procedures for runtime visual decision making. Each MMSkill couples a \textit{textual procedure}, which describes the reusable action pattern, with \textit{runtime state cards}, which encode when-to-use and when-not-to-use conditions, visible cues, verification cues, and available views, and \textit{multi-view keyframes}, which ground critical states through full-frame, focused, and optional before/after views. The resulting package is not a text instruction with illustrative images attached. It is a state-conditioned procedure whose visual evidence helps the agent decide when to follow, skip, or verify the procedure.

\begin{figure}[t]
    \centering
    \includegraphics[width=\textwidth]{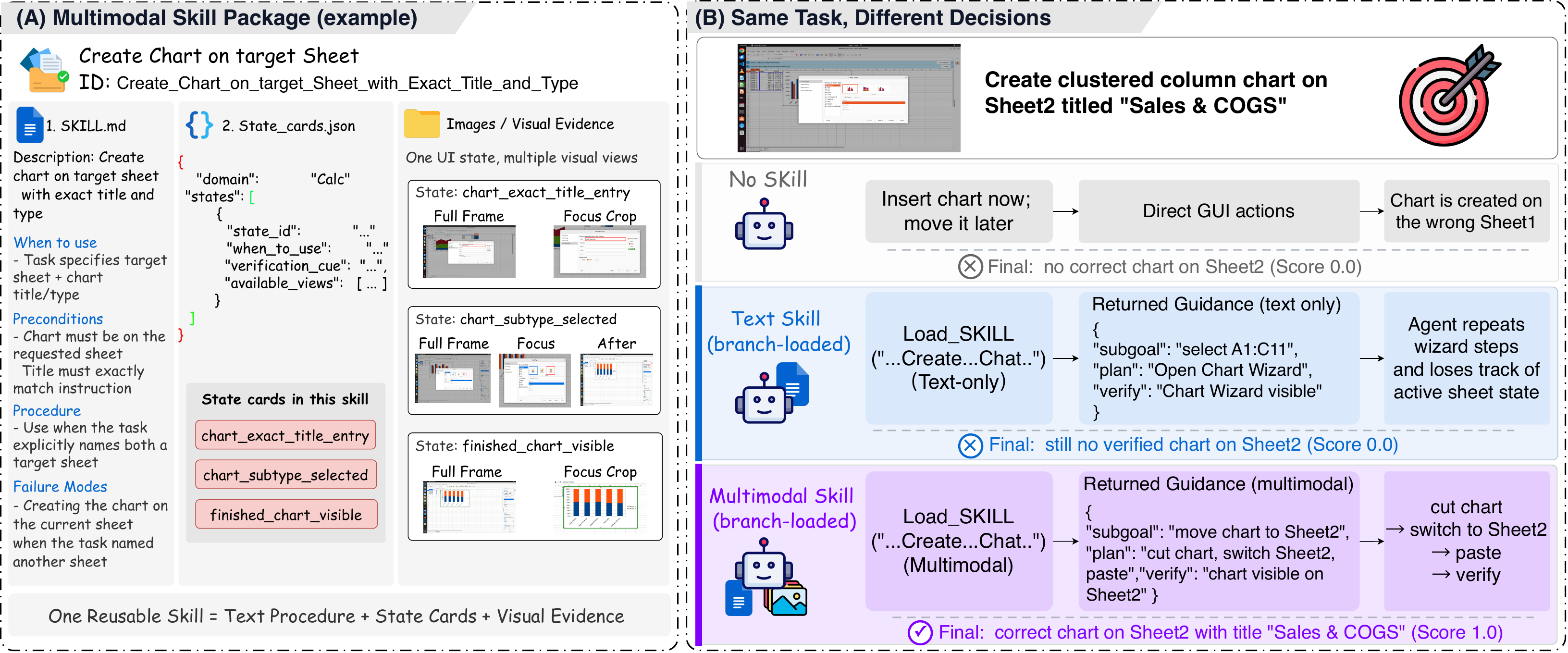}
    \caption{A concrete MMSkills example. A multimodal skill package combines a textual procedure, runtime state cards, and multi-view visual evidence. For the same chart-creation task, text-only guidance can miss the active sheet state, while branch-loaded MMSkills align skill evidence with the live screen and return state-aware guidance for the main agent.}
    \label{fig:intro-mmskill-example}
\end{figure}

To \textbf{generate} the multimodal skill package, we introduce an \textit{\textbf{automated trajectory-to-skill Generator}} built around an agentic, meta-skill-guided pipeline. This generation problem is substantially harder than text-skill extraction: while prior pipelines can often compress successful rollouts, failure analyses, or accumulated traces into reusable instructions or action abstractions \citep{zheng2025skillweaver,wang2026skillx,alzubi2026evoskill,ma2026skillclawletskillsevolve,xia2026skillrlevolvingagentsrecursive,li2026skillsbenchbenchmarkingagentskills}, generating MMSkills must also identify reusable visual states, select diagnostic frames, and bind each visual cue to the decision rule it supports. Our Generator operates on public trajectories that are \textbf{separate from evaluation tasks}: it groups related workflows, induces candidate procedures, merges overlapping candidates, grounds them in real non-test trajectory frames, and audits the resulting packages with reusable multimodal-skill-factory meta-skills. This process converts public interaction data into compact visual procedural knowledge without storing raw demonstrations as the skill.

For effective \textbf{utilization}, we introduce \textit{\textbf{branch loading}} to consult the multimodal skills without injecting the entire package into the main trajectory. Existing skill agents commonly insert retrieved skills directly into the main interaction context. This loading pattern becomes problematic for MMSkills: a single package may contain several state cards together with multi-view screenshots, so direct insertion creates substantial context pressure and makes reference images compete with the live observation. More importantly, the main agent can become visually anchored to superficially similar reference screenshots, planning around the skill example rather than the current environment. Branch loading addresses this issue as a multimodal form of progressive disclosure over skill evidence \citep{xu2026agentskills}. When the main agent considers a skill, it opens a temporary branch that selects the needed state cards and keyframe views, aligns them with the live screen or scene, and returns compact structured guidance with applicability judgments, subgoals, and next-step plans. The main trajectory receives distilled decision support rather than the full skill package, as illustrated by the example in Figure~\ref{fig:intro-mmskill-example}.

We evaluate MMSkills across GUI and game-based visual agent tasks, including OSWorld \citep{xie2024osworld}, macOSWorld \citep{yang2025macosworld}, VAB-Minecraft from VisualAgentBench \citep{liu2024visualagentbench}, and Super-Mario in LMGame-Bench \citep{hu2025lmgamebenchgoodllmsplaying}. Across frontier and smaller multimodal models, MMSkills improve performance over no-skill and text-only skill conditions, suggesting that external visual procedural knowledge complements model-internal priors.

Our main contributions are summarized as follows:
\begin{itemize}[leftmargin=*]
    \item To the best of our knowledge, we are the first to introduce the \textbf{multimodal skill package}, formulating reusable skills for general visual agents as multimodal procedural knowledge: compact, state-conditioned units that organize textual procedures, runtime state cards, and multi-view keyframes for visual decision making.
    \item We develop an agentic trajectory-to-skill \textbf{Generator} that turns public, non-evaluation trajectories into multimodal skill packages through workflow grouping, procedure induction, visual grounding, and meta-skill-guided auditing.
    \item We propose \textbf{branch loading}, a runtime mechanism that selects and aligns multimodal skill evidence in a temporary branch before returning structured decision support to the main agent.
    \item We demonstrate significant gains across GUI and game-based visual-agent benchmarks and multiple model families, showing that external multimodal procedural knowledge complements model-internal priors.
\end{itemize}


\section{Methods}
\label{sec:methods}

\newcommand{\mfield}[1]{\text{\normalfont\ttfamily #1}}

\subsection{Overview}
\label{sec:method-overview}

\begin{figure}[t]
    \centering
    \includegraphics[width=\textwidth]{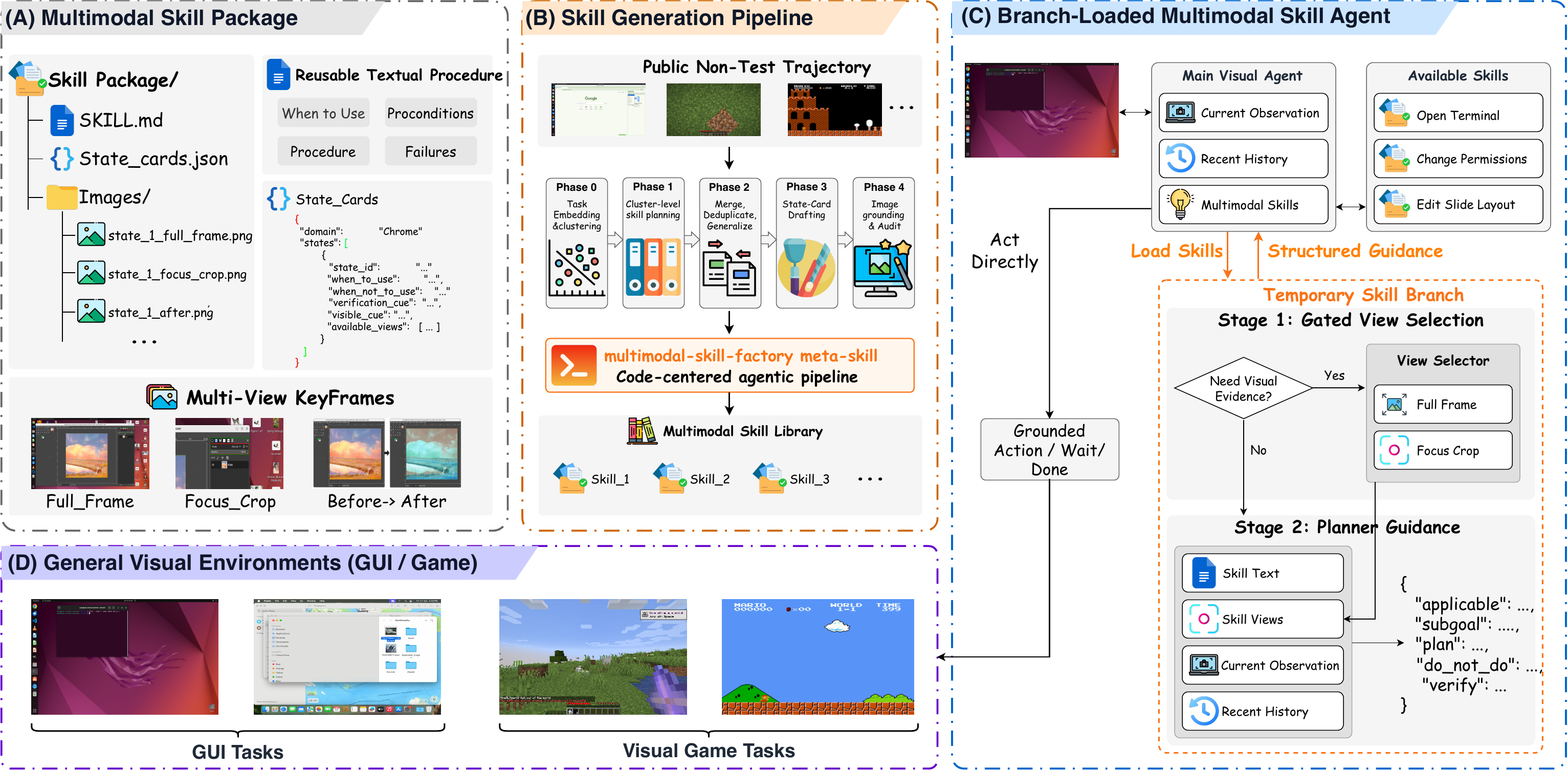}
    \caption{Overview of the MMSkills framework. A multimodal skill package stores a reusable textual procedure, runtime state cards, and multi-view keyframes. A meta-skill-guided Generator converts public non-test trajectories into a reusable multimodal skill library. At inference time, the main visual agent uses branch loading to inspect selected skill evidence in a temporary branch and receives compact structured guidance before acting.}
    \label{fig:method-overview}
\end{figure}

MMSkills are designed around three components: a \emph{multimodal skill package} that stores reusable visual procedural knowledge, a \emph{Skill Generation pipeline} that constructs such packages from public trajectories, and a \emph{branch-loaded multimodal skill agent} that isolates skill-environment grounding in a temporary branch and returns distilled decision support to the main trajectory at inference time. Figure~\ref{fig:method-overview} gives the system overview.

At a high level, the Generator maps non-evaluation trajectories $\mathcal{T}=\{\tau_i\}$ into a multimodal skill library $\mathcal{M}=\{M_i\}_{i=1}^{N}$. Before an episode begins, the runtime agent pre-recalls a task-level candidate set $\mathcal{C}_I \subset \mathcal{M}$ from the instruction $I$ and compact skill descriptors. During execution, the main agent observes the current visual observation $O_t$, maintains a short history $H_t$, and either acts directly or consults a temporary skill branch for some $M_t\in\mathcal{C}_I$:
\begin{equation}
\begin{aligned}
\text{direct}: \quad
    & A_t = \pi_{\text{main}}(O_t,H_t,\mathcal{C}_I),\\
\text{branch}: \quad
    & G_t = \text{Branch}(O_t,H_t,M_t),\quad
      A_t = \pi_{\text{main}}(O_t,H_t,\mathcal{C}_I,G_t).
\end{aligned}
\label{eq:runtime-modes}
\end{equation}
The branch output is a structured guidance tuple
\begin{equation}
    G_t = (\text{applicable}_t,\text{subgoal}_t,\text{plan}_t,\text{do\_not\_do}_t,\text{verify}_t),
    \label{eq:branch-summary}
\end{equation}
where the fields respectively give the applicability judgment, local subgoal, skill-conditioned plan, negative constraints, and visual verification check. The main agent uses $G_t$ as decision support, while executable action grounding remains tied to the live observation.

\subsection{Multimodal Skill Package}
\label{sec:method-package}

We represent each MMSkill as a state-conditioned procedure package
\begin{equation}
    M=(D,P,S,K),
    \label{eq:mmskill}
\end{equation}
where $D$ is a compact descriptor, $P$ is a reusable textual procedure, $S=\{S_j\}_{j=1}^{m}$ is a set of runtime state cards, and $K=\{K_j\}_{j=1}^{m}$ is a set of keyframe bundles aligned with those cards. Each pair $(S_j,K_j)$ corresponds to one decision-relevant procedural state. The procedure specifies the reusable workflow; the state card specifies when the workflow is valid or invalid; and the keyframes make the state visually recognizable at runtime.

A runtime state card is an agent-facing state node rather than an image caption. It links a point in the procedure to when-to-use conditions, when-not-to-use conditions, visible cues, verification cues, and available views:
\begin{equation}
\begin{split}
S_j = (&
\text{when\_to\_use}_j,
\text{when\_not\_to\_use}_j,
\text{visible\_cues}_j,\\
&
\text{verification\_cue}_j,
\mathcal{V}_j),
\qquad \mathcal{V}_j=\text{available\_views}_j .
\end{split}
\label{eq:state-card}
\end{equation}
The first two fields define when the state should be followed or skipped, $\mfield{visible\_cues}_j$ states what evidence to inspect, $\mfield{verification\_cue}_j$ defines the progress or completion check, and $\mathcal{V}_j$ lists which views may be loaded. This schema makes the skill useful for decision making: the agent can decide whether to follow, skip, or verify the procedure.

Each key state is grounded by a small multi-view bundle. Let
\begin{equation}
    \mathcal{V}=\{\text{full\_frame},\text{focus\_crop},\text{before},\text{after}\}.
\end{equation}
Then
\begin{equation}
    K_j=\{K_j^{v}:v\in\mathcal{V}_j,\ v\in\mathcal{V}\}.
    \label{eq:keyframe-bundle}
\end{equation}
The full-frame view preserves global context, the focus crop localizes the visual cue, and optional before/after views expose useful transitions. These images are reference evidence, not coordinates to copy. Under this representation, a text-only skill is the degenerate package $(D,P,\emptyset,\emptyset)$; MMSkills extend it by binding procedure, decision conditions, and visual evidence into one reusable unit.

\subsection{Skill Generator from Public Trajectories}
\label{sec:method-generator}

We build MMSkills from public interaction trajectories that are separate from evaluation tasks. A trajectory is
\begin{equation}
    \tau_i=(I_i,O_{i,1:T_i},A_{i,1:T_i}),
\end{equation}
where $I_i$ is the task instruction, $O_{i,t}$ are visual observations, $A_{i,t}$ are executed actions. The Generator is controlled by a reusable multimodal-skill-factory meta-skill $\mathcal{F}$:
\begin{equation}
    \mathcal{G}_{\mathcal{F}}:\mathcal{T}_d\mapsto\mathcal{M}_d,
    \label{eq:meta-generator}
\end{equation}
where $\mathcal{T}_d$ is the public trajectory pool for domain $d$ and $\mathcal{M}_d$ is the generated domain skill library. The pipeline comprises five stages:
\begin{equation}
\begin{aligned}
    \mathcal{T}_d
    &\xrightarrow{\text{Phase 0: embed+cluster}} \mathcal{C}_d
    \xrightarrow{\text{Phase 1: cluster plan}} \mathcal{A}_d
    \xrightarrow{\text{Phase 2: merge}} \mathcal{R}_d \\
    &\xrightarrow{\text{Phase 3: text draft}} \widehat{\mathcal{M}}_d
    \xrightarrow{\text{Phase 4: image ground+audit}} \mathcal{M}_d .
\end{aligned}
\label{eq:generator-pipeline}
\end{equation}

\begin{itemize}[leftmargin=*]
    \item \textbf{Phase 0: task embedding and clustering.} The pipeline embeds task instructions and trajectory metadata, then groups a broad domain into semantically focused clusters $\mathcal{C}_d$.
    \item \textbf{Phase 1: cluster-level skill planning.} For each cluster, an LLM-based agent proposes atomic skills with workflow boundaries, completion conditions, and covered task ids, producing a domain planning table $\mathcal{A}_d$.
    \item \textbf{Phase 2: skill merging.} Cluster-level plans are deduplicated, merged, and generalized into merged skill specifications $\mathcal{R}_d$, while overly broad umbrella skills are rejected.
    \item \textbf{Phase 3: text-first drafting.} Without reading images, the Generator selects reference tasks and drafts the descriptor $D$, textual procedure $P$, and planned state cards, yielding $\widehat{\mathcal{M}}_d$.
    \item \textbf{Phase 4: image grounding and audit.} The Generator reads selected keyframes, grounds focus regions, constructs multi-view bundles, and audits the final packages.
\end{itemize}

For a merged skill $r\in\mathcal{R}_d$, finalization is written as
\begin{equation}
    \widehat{M}_r=(D_r,P_r,\widehat{S}_r,\widehat{K}_r)
    \xrightarrow{\text{ground+audit}}
    M_r=(D_r,P_r,S_r,K_r).
\end{equation}
The visual grounding policy is conservative: views are added only for state recognition, transition comparison, or completion verification, so the skill stores diagnostic states rather than replaying demonstrations. The meta-skill $\mathcal{F}$ supplies reusable scripts, schemas, and quality gates for the LLM-based Generator, while external services are limited to bounded support steps such as embedding/clustering and grounding.

\subsection{Branch-loaded Multimodal Skills Agent}
\label{sec:method-branch}

Most skill-using agents load a retrieved skill directly into the main interaction context. For short text skills, this is reasonable: the skill is read as an additional instruction alongside the observation. For MMSkills, direct loading is brittle because state cards, multi-view keyframes, and transition examples add substantial context pressure, and irrelevant reference views can anchor the agent away from the live environment. Figure~\ref{fig:method-overview}(C) illustrates the branch-loaded alternative, which moves skill-environment grounding out of the main trajectory.

\textbf{Stage 1: gated view selection.} Suppose the main agent calls $M_t=(D_t,P_t,S_t,K_t)\in\mathcal{C}_I$. The branch first selects which state cards and view types are relevant to the live observation:
\begin{equation}
    (J_t,R_t)=\text{SelectViews}(O_t,H_{t-1},P_t,S_t),
    \qquad
    V_t=\{K_j^v:j\in J_t,\ v\in R_{t,j}\},
    \label{eq:branch-stage1}
\end{equation}
where $J_t$ indexes selected state cards and $R_{t,j}\subseteq\mathcal{V}_j$ selects views for state $j$. The selector reads the live observation, recent history, textual procedure, and state-card descriptions before loading images. If text and state cards are sufficient, $R_{t,j}$ may be empty.

\textbf{Stage 2: branch planning.} The branch then aligns the selected evidence with the live state and returns structured guidance:
\begin{equation}
    G_t=\text{PlanBranch}(O_t,H_{t-1},P_t,\{S_j:j\in J_t\},V_t),
    \label{eq:branch-stage2}
\end{equation}
where $G_t$ follows Eq.~\ref{eq:branch-summary}. The main agent does not execute $G_t$ mechanically; it uses $G_t$ as an intermediate planning signal and still chooses a grounded action from the live screenshot. This preserves procedural guidance without allowing reference images to override the current observation. Appendix~\ref{app:branch-loaded-algorithm} gives the full runtime loop in Algorithm~\ref{alg:branch-loaded-agent}, and Appendix~\ref{app:mmskillagent-prompts} reports the prompt templates used by the main agent and the two branch stages.


\section{Experiments}
\label{sec:experiments}

We evaluate whether MMSkills provide useful external procedural knowledge for visual agents. The experiments are organized around four research questions:

\begin{itemize}[leftmargin=*]
    \item \textbf{RQ1: Overall performance on GUI and game tasks.} Do MMSkills improve visual agents across realistic desktop environments and open-ended visual game tasks?
    \item \textbf{RQ2: Ablations of skill content and branch loading.} Which parts of MMSkills matter, and how do branch loading and view selection affect multimodal skill use?
    \item \textbf{RQ3: Skill usage and interaction dynamics.} How often are MMSkills invoked, how do they affect interaction length, and which visual views are selected at runtime?
    \item \textbf{RQ4: Behavioral shift analysis.} How do MMSkills change the agent's low-level action patterns beyond final success rate?
\end{itemize}

\definecolor{mmskillrowpurple}{RGB}{244,238,255}
\renewcommand{\mmskillhl}{\cellcolor{mmskillrowpurple}}

\begin{table}[!t]
\centering
\caption{OSWorld application-level success rates. All entries are percentages. ``Calc'', ``Impress'', and ``Writer'' denote LibreOffice applications.}
\label{tab:osworld-domain-results}
\scriptsize
\setlength{\tabcolsep}{2.4pt}
\resizebox{\textwidth}{!}{%
\begin{tabular}{ccccccccccccc}
\toprule
Base model & Skill condition & Chrome & GIMP & Calc & Impress & Writer & Multi-app & OS & Mail & VLC & VS Code & Overall \\
\midrule
\multirow{3}{*}{\parbox{1.9cm}{\centering Gemini 3.1 Pro}}
 & No skill & 53.47 & 34.62 & \textbf{57.45} & 40.43 & 47.82 & \textbf{31.97} & 54.17 & 40.00 & 35.29 & 56.52 & 44.08 \\
 & Text-only & 44.35 & 34.62 & 38.30 & 40.34 & 56.52 & 22.38 & \textbf{70.83} & \textbf{66.67} & 41.18 & 56.52 & 40.76 \\
 & \mmskillhl \textbf{MMSkills} & \mmskillhl \textbf{59.91} & \mmskillhl \textbf{50.00} & \mmskillhl 53.19 & \mmskillhl \textbf{53.19} & \mmskillhl \textbf{60.86} & \mmskillhl 24.11 & \mmskillhl \textbf{70.83} & \mmskillhl \textbf{66.67} & \mmskillhl \textbf{70.59} & \mmskillhl \textbf{65.22} & \mmskillhl \textbf{50.11} \\
\midrule
\multirow{3}{*}{\parbox{1.9cm}{\centering Gemini 3 Flash}}
 & No skill & 37.78 & 50.00 & 38.30 & 29.73 & 52.17 & 21.51 & 54.17 & 66.67 & 52.39 & 47.83 & 36.65 \\
 & Text-only & 51.02 & 23.08 & 38.30 & 34.00 & 56.52 & 19.16 & 54.17 & 60.00 & 58.82 & 52.17 & 40.27 \\
 & \mmskillhl \textbf{MMSkills} & \mmskillhl \textbf{55.37} & \mmskillhl 42.31 & \mmskillhl \textbf{53.19} & \mmskillhl \textbf{40.34} & \mmskillhl \textbf{56.52} & \mmskillhl \textbf{30.98} & \mmskillhl \textbf{75.00} & \mmskillhl \textbf{66.67} & \mmskillhl 52.94 & \mmskillhl \textbf{60.87} & \mmskillhl \textbf{47.97} \\
\midrule
\multirow{3}{*}{\parbox{1.9cm}{\centering Qwen3-VL-235B}}
 & No skill & 15.56 & 38.46 & 17.02 & 25.53 & 43.48 & 9.48 & 25.00 & 26.67 & 17.65 & 34.78 & 21.34 \\
 & Text-only & 42.22 & 50.00 & 10.64 & 21.31 & 34.78 & 14.86 & 33.33 & 60.00 & 35.29 & 47.83 & 28.57 \\
 & \mmskillhl \textbf{MMSkills} & \mmskillhl \textbf{59.91} & \mmskillhl \textbf{69.23} & \mmskillhl \textbf{23.40} & \mmskillhl \textbf{32.01} & \mmskillhl \textbf{47.82} & \mmskillhl \textbf{19.35} & \mmskillhl \textbf{41.67} & \mmskillhl \textbf{73.33} & \mmskillhl \textbf{41.18} & \mmskillhl \textbf{56.52} & \mmskillhl \textbf{39.17} \\
\midrule
\multirow{3}{*}{\parbox{1.9cm}{\centering GLM-5V}}
 & No skill & 37.78 & 19.23 & 21.28 & 29.70 & 26.08 & 18.70 & 54.17 & 53.33 & 11.76 & 47.83 & 28.71 \\
 & Text-only & \textbf{53.24} & \textbf{53.85} & \textbf{31.91} & \textbf{31.98} & \textbf{52.17} & 20.24 & 20.83 & \textbf{46.67} & \textbf{35.29} & \textbf{65.22} & 36.61 \\
 & \mmskillhl \textbf{MMSkills} & \mmskillhl 51.02 & \mmskillhl \textbf{53.85} & \mmskillhl \textbf{31.91} & \mmskillhl 31.83 & \mmskillhl 43.47 & \mmskillhl \textbf{22.26} & \mmskillhl \textbf{66.67} & \mmskillhl 40.00 & \mmskillhl 23.53 & \mmskillhl \textbf{65.22} & \mmskillhl \textbf{38.51} \\
\midrule
\multirow{3}{*}{\parbox{1.9cm}{\centering Kimi-K2.6}}
 & No skill & 51.02 & 34.62 & 34.04 & 35.32 & 30.43 & 14.86 & 54.17 & 66.67 & 32.60 & 52.17 & 34.98 \\
 & Text-only & \textbf{57.69} & 40.00 & \textbf{40.43} & 36.14 & 17.38 & 22.38 & 62.50 & 53.33 & \textbf{58.82} & 43.48 & 39.66 \\
 & \mmskillhl \textbf{MMSkills} & \mmskillhl \textbf{57.69} & \mmskillhl \textbf{42.31} & \mmskillhl \textbf{40.43} & \mmskillhl \textbf{48.92} & \mmskillhl \textbf{60.86} & \mmskillhl \textbf{23.40} & \mmskillhl \textbf{79.17} & \mmskillhl \textbf{73.33} & \mmskillhl 41.18 & \mmskillhl \textbf{69.57} & \mmskillhl \textbf{46.59} \\
\midrule
\multirow{3}{*}{\parbox{1.9cm}{\centering Qwen3-VL-8B-\\Instruct}}
 & No skill & 15.47 & 7.69 & 2.13 & 8.59 & 4.34 & 7.33 & 25.00 & 13.33 & 29.41 & 17.39 & 10.78 \\
 & Text-only & 19.91 & 11.54 & 6.38 & 16.99 & \textbf{17.39} & 7.33 & 16.67 & 33.33 & 17.65 & 34.78 & 14.93 \\
 & \mmskillhl \textbf{MMSkills} & \mmskillhl \textbf{39.91} & \mmskillhl \textbf{42.31} & \mmskillhl \textbf{8.51} & \mmskillhl \textbf{23.37} & \mmskillhl \textbf{17.39} & \mmskillhl \textbf{13.43} & \mmskillhl \textbf{25.00} & \mmskillhl \textbf{60.00} & \mmskillhl \textbf{29.41} & \mmskillhl \textbf{47.83} & \mmskillhl \textbf{25.40} \\
\bottomrule
\end{tabular}}
\vspace{-0.15em}
\parbox{\textwidth}{\scriptsize\emph{Note:} Due to the substantially higher inference cost and wall-clock time of Gemini 3.1 Pro and Kimi-K2.6, we report their full three-condition results only on OSWorld.}
\end{table}

\begin{table}[!t]
\centering
\caption{Auxiliary GUI and game-based visual-agent results. macOSWorld reports domain-level and overall success rates; VAB-Minecraft reports success rate and average score; Super Mario Bros reports total performance and total reward.}
\label{tab:auxiliary-results}
\scriptsize
\setlength{\tabcolsep}{2.2pt}
\resizebox{\textwidth}{!}{%
\begin{tabular}{cccccccccccc}
\toprule
& & \multicolumn{6}{c}{macOSWorld} & \multicolumn{2}{c}{VAB-Minecraft} & \multicolumn{2}{c}{Super Mario Bros} \\
\cmidrule(lr){3-8}\cmidrule(lr){9-10}\cmidrule(lr){11-12}
Base model & Skill condition & File & Media & Prod. & Sys/IF & Apps & Overall & Success & Avg. score & Total perf. & Total reward \\
\midrule
\multirow{3}{*}{\parbox{1.9cm}{\centering Gemini 3 Flash}}
 & No skill & 41.38 & 33.33 & 60.00 & 62.07 & 55.79 & 55.94 & 67.24 & 0.7462 & 411.00 & 766.67 \\
 & Text-only & 31.03 & 25.00 & 62.86 & \textbf{75.86} & 55.26 & 53.85 & 68.96 & 0.7541 & 548.00 & 912.00 \\
 & \mmskillhl \textbf{MMSkills} & \mmskillhl \textbf{58.62} & \mmskillhl \textbf{50.00} & \mmskillhl \textbf{77.14} & \mmskillhl 65.52 & \mmskillhl \textbf{65.73} & \mmskillhl \textbf{65.73} & \mmskillhl \textbf{73.28} & \mmskillhl \textbf{0.7884} & \mmskillhl \textbf{624.00} & \mmskillhl \textbf{1081.33} \\
\midrule
\multirow{3}{*}{\parbox{1.9cm}{\centering Qwen3-VL-235B}}
 & No skill & 31.03 & \textbf{58.33} & 51.43 & 58.62 & 44.74 & 47.55 & 52.59 & 0.6308 & 454.50 & 955.50 \\
 & Text-only & 34.48 & 33.33 & 37.14 & 51.72 & 52.63 & 43.36 & 55.17 & 0.6634 & 610.50 & 1138.25 \\
 & \mmskillhl \textbf{MMSkills} & \mmskillhl \textbf{37.93} & \mmskillhl 33.33 & \mmskillhl \textbf{54.29} & \mmskillhl \textbf{62.07} & \mmskillhl \textbf{57.89} & \mmskillhl \textbf{51.75} & \mmskillhl \textbf{62.07} & \mmskillhl \textbf{0.7114} & \mmskillhl \textbf{788.00} & \mmskillhl \textbf{1514.25} \\
\midrule
\multirow{3}{*}{\parbox{1.9cm}{\centering GLM-5V}}
 & No skill & 24.14 & 16.67 & 40.00 & 41.38 & 39.47 & 34.97 & 56.03 & 0.6701 & 612.75 & 1191.50 \\
 & Text-only & 31.03 & \textbf{66.67} & \textbf{62.86} & \textbf{58.62} & 47.37 & \textbf{51.75} & 61.20 & 0.6938 & 794.50 & 1218.00 \\
 & \mmskillhl \textbf{MMSkills} & \mmskillhl \textbf{44.83} & \mmskillhl \textbf{66.67} & \mmskillhl 48.57 & \mmskillhl \textbf{58.62} & \mmskillhl \textbf{50.00} & \mmskillhl \textbf{51.75} & \mmskillhl \textbf{68.10} & \mmskillhl \textbf{0.7495} & \mmskillhl \textbf{950.50} & \mmskillhl \textbf{1384.50} \\
\midrule
\multirow{3}{*}{\parbox{1.9cm}{\centering Qwen3-VL-8B-\\Instruct}}
 & No skill & \textbf{10.34} & 0.00 & \textbf{14.29} & \textbf{3.45} & 0.00 & \textbf{6.29} & 23.28 & 0.3017 & 415.25 & 928.75 \\
 & Text-only & 0.00 & \textbf{8.33} & 2.86 & \textbf{3.45} & \textbf{10.53} & 4.90 & 29.31 & 0.3754 & 596.50 & 997.25 \\
 & \mmskillhl \textbf{MMSkills} & \mmskillhl 6.90 & \mmskillhl \textbf{8.33} & \mmskillhl 8.57 & \mmskillhl \textbf{3.45} & \mmskillhl 5.26 & \mmskillhl \textbf{6.29} & \mmskillhl \textbf{38.79} & \mmskillhl \textbf{0.4668} & \mmskillhl \textbf{764.00} & \mmskillhl \textbf{1128.75} \\
\bottomrule
\end{tabular}}
\end{table}

\renewcommand{\mmskillhl}{\cellcolor{mmskillrow}}

\subsection{Experimental Setup}
\label{sec:experiments-setup}

In all settings, agents plan from visual observations, namely desktop or game screenshots. We evaluate on OSWorld \citep{xie2024osworld}, macOSWorld \citep{yang2025macosworld}, VAB-Minecraft from VisualAgentBench \citep{liu2024visualagentbench}, and Super Mario Bros from LMGame-Bench \citep{hu2025lmgamebenchgoodllmsplaying}, covering both realistic GUI tasks and open visual game environments. Detailed benchmark descriptions and test-case distributions are illustrated in Appendix~\ref{app:benchmark-statistics}; implementation details, evaluation protocols, model choices, and runtime variants are given in Appendix~\ref{app:experiment-details}.

\textbf{All skills are extracted from non-test data}. We evaluate frontier and smaller multimodal models and compare \emph{no-skill}, \emph{text-only skill}, and \emph{MMSkills} conditions, with direct-loading variants studied in the ablations. Dataset-specific skill sources, source statistics, and skill-package distributions are provided in Appendix~\ref{app:skill-source-statistics}.

\subsection{RQ1: Overall Performance on GUI and Game Tasks}
\label{sec:experiments-desktop}

Table~\ref{tab:osworld-domain-results} reports OSWorld application-level success rates, and Table~\ref{tab:auxiliary-results} reports the auxiliary GUI and game results. \textbf{MMSkills improve OSWorld overall performance across all evaluated model families.} Overall success increases for Gemini 3.1 Pro ($44.08\%\!\to\!50.11\%$), Gemini 3 Flash ($36.65\%\!\to\!47.97\%$), Qwen3-VL-235B ($21.34\%\!\to\!39.17\%$), GLM-5V, and Kimi-K2.6. Text-only skills help but are less stable across domains, suggesting that procedures alone are insufficient when skill use depends on visual state matching. \textbf{External multimodal procedural knowledge is especially valuable for weaker visual agents.} For Qwen3-VL-8B-Instruct, MMSkills raise OSWorld from $10.78\%$ to $25.40\%$ and VAB-Minecraft from $23.28\%$ to $38.79\%$, indicating that explicit visual procedural knowledge can compensate for limited model-internal priors.

\textbf{The gains transfer beyond Ubuntu desktop tasks.} On macOSWorld, MMSkills improve the completed large-model runs, including Gemini 3 Flash and GLM-5V, while VAB-Minecraft shows consistent gains in both success rate and average score across all evaluated models. Super Mario Bros follows the same pattern in the completed runs, with higher total performance and reward under MMSkills. These results indicate that MMSkills are not specialized to a single GUI benchmark; the same state-conditioned skill format helps in visually grounded game settings where recurring states and action strategies can be reused.

\subsection{RQ2: Ablations of Skill Content and Branch Loading}
\label{sec:experiments-ablation}

\begin{figure}[t]
    \centering
    \includegraphics[width=\textwidth]{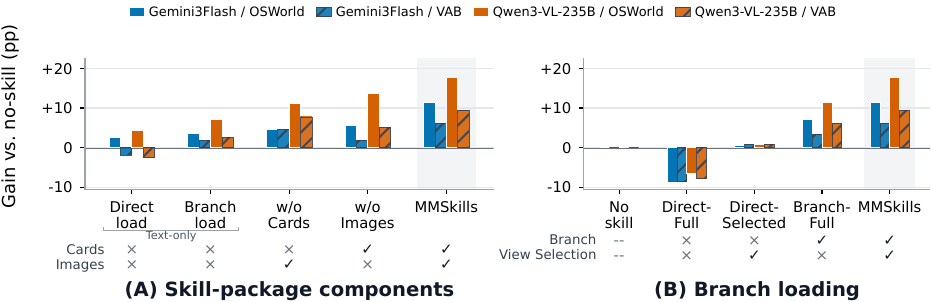}
    \caption{Ablation results for MMSkills components and branch loading. Bars report percentage-point gains over the no-skill baseline. Panel (A) removes runtime state cards or visual keyframes from the skill package. Panel (B) compares direct loading with branch loading and with or without view selection.}
    \label{fig:ablation-results}
\end{figure}

Figure~\ref{fig:ablation-results} combines the skill-content and branch-loading ablations. Unless otherwise stated, skill variants use the branch-loaded agent; the main exception is \emph{Direct load}, which inserts skill content into the main context. For skill content, we compare text-only skills, MMSkills without state cards, MMSkills without images, and the complete MMSkills package. \textbf{State cards and multi-view visual evidence both improve skill utility.} Text-only branch loading already improves over the no-skill baseline, but the complete MMSkills package is consistently stronger. Removing state cards weakens the agent's ability to distinguish relevant runtime states, while removing images preserves decision rules but removes visual grounding evidence. Both removals reduce performance on OSWorld and VAB-Minecraft, confirming that state cards and keyframes play complementary roles: one supports state discrimination, and the other helps the agent recognize the corresponding visual evidence. \textbf{Branch loading helps even for text-only skills.} The branch-loaded text-only variant is stronger than direct text loading in most model--benchmark pairs, indicating that the temporary branch improves skill interpretation even before multimodal evidence is introduced.

For branch loading, we ablate whether skill evidence is inspected in a temporary branch and whether Stage-1 view selection filters state cards and keyframes. \textbf{Branch loading and view selection address different failure modes.} Direct-full loading hurts performance because unfiltered images and state descriptions pollute the main context; view selection alone reduces this damage but stays near baseline. Branch loading already gives clear gains, and the full two-stage design performs best, indicating that separated evidence inspection and filtered visual evidence are both necessary.

\subsection{RQ3: Skill Usage and Interaction Dynamics}
\label{sec:experiments-analysis}

Table~\ref{tab:skill-usage-analysis} analyzes when and how agents call skills. \textbf{MMSkills are invoked more often than text-only skills.} Invocation coverage increases on both OSWorld and VAB-Minecraft for Gemini 3 Flash and Qwen3-VL-235B, with the largest OSWorld change rising from $37.50\%$ to $65.28\%$ for Qwen3-VL-235B. This suggests that multimodal skills make external knowledge easier to recognize as relevant: state cards expose when-to-use and when-not-to-use conditions, and visual cues help the agent detect when its current observation matches a reusable procedural state.

\renewcommand{\mmskillhl}{\cellcolor{mmskillrowpurple}}

\begin{table}[t]
\centering
\caption{Skill invocation, interaction length, and selected views. ``Invoked'' is the percentage of cases with at least one skill call, and ``Step $\Delta$'' is relative to the no-skill baseline.}
\label{tab:skill-usage-analysis}
\scriptsize
\setlength{\tabcolsep}{4pt}
\resizebox{\textwidth}{!}{%
\begin{tabular}{cccccccc}
\toprule
Benchmark & Model & Skill condition & Invoked (\%) & Calls/case & Steps & Step $\Delta$ & Views (Full/Focus/Before/After) \\
\midrule
\multirow{6}{*}{OSWorld}
 & \multirow{3}{*}{Gemini 3 Flash}
 & No skill & -- & -- & 13.11 & 0.00 & -- \\
 & & Text-only & 41.11 & 0.7139 & 15.64 & +2.53 & -- \\
 & & \mmskillhl \textbf{MMSkills} & \mmskillhl \textbf{62.50} & \mmskillhl \textbf{0.9556} & \mmskillhl \textbf{11.86} & \mmskillhl \textbf{-1.25} & \mmskillhl 79/241/8/24 \\
\cmidrule(lr){2-8}
 & \multirow{3}{*}{Qwen3-VL-235B}
 & No skill & -- & -- & 15.22 & 0.00 & -- \\
 & & Text-only & 37.50 & 0.4917 & 13.34 & -1.88 & -- \\
 & & \mmskillhl \textbf{MMSkills} & \mmskillhl \textbf{65.28} & \mmskillhl \textbf{0.9222} & \mmskillhl \textbf{9.87} & \mmskillhl \textbf{-5.35} & \mmskillhl 40/27/17/13 \\
\midrule
\multirow{6}{*}{VAB-Minecraft}
 & \multirow{3}{*}{Gemini 3 Flash}
 & No skill & -- & -- & 16.92 & 0.00 & -- \\
 & & Text-only & 68.97 & 1.8706 & 17.30 & +0.38 & -- \\
 & & \mmskillhl \textbf{MMSkills} & \mmskillhl \textbf{81.90} & \mmskillhl \textbf{2.4310} & \mmskillhl \textbf{13.75} & \mmskillhl \textbf{-3.17} & \mmskillhl 105/205/15/12 \\
\cmidrule(lr){2-8}
 & \multirow{3}{*}{Qwen3-VL-235B}
 & No skill & -- & -- & 34.74 & 0.00 & -- \\
 & & Text-only & 54.31 & 1.5776 & 31.36 & -3.38 & -- \\
 & & \mmskillhl \textbf{MMSkills} & \mmskillhl \textbf{64.66} & \mmskillhl \textbf{2.3534} & \mmskillhl \textbf{27.07} & \mmskillhl \textbf{-7.67} & \mmskillhl 98/196/13/10 \\
\bottomrule
\end{tabular}}
\end{table}

\renewcommand{\mmskillhl}{\cellcolor{mmskillrow}}

\textbf{MMSkills shorten trajectories rather than merely adding extra consultation.} Text-only skills can add overhead when they provide procedural hints without visual grounding, but MMSkills reduce average steps in every setting, with the largest reductions appearing for Qwen3-VL-235B. These reductions indicate that multimodal skills help agents find shorter task-solving paths and avoid unnecessary exploration or repeated low-value actions. \textbf{Focus crops dominate selected visual evidence.} The branch does not load all views uniformly: focus crops are selected most frequently in three of four settings, while full-frame, before, and after views provide global context, transition evidence, and completion references when local crops alone are insufficient.

\subsection{RQ4: Behavioral Shift Analysis}
\label{sec:experiments-behavior}

\begin{figure}[t]
    \centering
    \includegraphics[width=\textwidth]{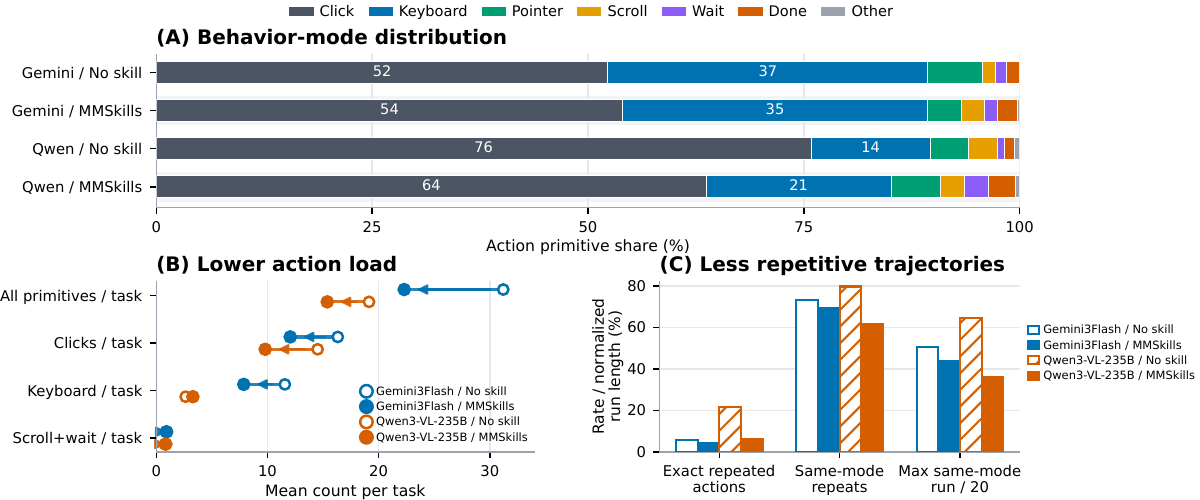}
    \caption{Behavioral shifts induced by MMSkills on OSWorld. Panel (A) reports the distribution of executed action primitives. Panel (B) compares the average number of low-level primitives per task. Panel (C) measures repetitive behavior through exact repeated actions, repeated action modes, and the longest same-mode run normalized by the 20-step budget.}
    \label{fig:behavior-shift}
\end{figure}

Figure~\ref{fig:behavior-shift} shows that the effect of MMSkills is not merely a success-rate gain. \textbf{MMSkills reduce low-level action load.} Gemini 3 Flash uses substantially fewer primitives per task, and Qwen3-VL-235B shows a similar reduction, especially in click actions. This supports the view that multimodal state cards and visual evidence constrain the agent's search space: the agent performs fewer exploratory GUI operations before reaching a useful state. \textbf{The behavioral shift is strongest for Qwen3-VL-235B.} Its click share drops from $75.8\%$ to $63.7\%$, while keyboard and DONE actions increase, suggesting that MMSkills help click-heavy agents move toward more structured input and stronger completion judgments.

\textbf{MMSkills suppress repetitive trajectories and improve completion awareness.} The effect is clearest for Qwen3-VL-235B: exact repeated actions fall from $21.8\%$ to $6.2\%$, and the longest same-mode run decreases substantially. Gemini 3 Flash shows the same direction of change, though from a stronger baseline. MMSkills also increase DONE behavior for both models, indicating that state cards and verification cues help agents decide not only what to do next, but also when the task is complete. Overall, MMSkills reshape agent behavior from exploratory trial-and-error toward grounded, state-aware execution; Appendix~\ref{app:additional-behavior-analysis} provides the GLM-5V and Kimi-K2.6 analysis.


\section{Related Work}
\label{sec:related-work}

\paragraph{Skills for agents.}
Skill reuse has roots in temporal abstraction and motor primitives \citep{sutton1999options,ijspeert2013dmp}, and recent LLM agents store reusable behavior as language, code, APIs, or learned libraries \citep{ichter2022saycan,liang2023codepolicies,yao2023react,shinn2023reflexion,wang2023voyager,zheng2025skillweaver,chen2026cuaskill,wang2026skillx,alzubi2026evoskill,ma2026skillclawletskillsevolve,xia2026skillrlevolvingagentsrecursive}. A complementary line treats accumulated experience as long-term agent memory \citep{park2023generativeagentsinteractivesimulacra,packer2024memgptllmsoperatingsystems}, while surveys and benchmarks evaluate skill relevance, selection, and safety \citep{xu2026agentskills,li2026skillsbenchbenchmarkingagentskills,wang2026skilltesterbenchmarkingutilitysecurity,liu2026agenticskillsworkwild}. MMSkills follows this modular view but stores state-conditioned multimodal packages and uses branch loading instead of inserting full skill memory; Appendix~\ref{app:related-work} expands the discussion.

\paragraph{Visual agents.}
Visual-agent benchmarks span web, mobile, desktop, and embodied environments \citep{deng2023mind2web,zhou2024webarena,koh2024visualwebarena,he2024webvoyager,rawles2025androidworld,xie2024osworld,yang2025macosworld,liu2024visualagentbench}, and model and framework work improves screenshot grounding and GUI control \citep{cheng2024seeclick,wu2024osatlas,qin2025uitars,agashe2024agents,hong2024cogagentvisuallanguagemodel,zheng2024gpt4visiongeneralistwebagent,zhang2023appagentmultimodalagentssmartphone,lu2024omniparserpurevisionbased}. Dedicated grounding benchmarks measure how reliably models localize UI elements from instructions \citep{li2025screenspotproguigroundingprofessional,gou2025navigatingdigitalworldhumans,wang2025mmbenchguihierarchicalmultiplatformevaluation,xu2025deskvisionlargescaledesktop}. MMSkills builds on these capabilities but operates higher: it tells the agent which procedural state matters and what visual evidence confirms it.

Closest to our work, Mirage-1 introduces hierarchical multimodal skills, XSkill extracts skills from visually grounded experience, and CUA-Skill represents computer-use skills as parameterized procedures and execution graphs \citep{xie2025mirage,jiang2026xskillcontinuallearningexperience,chen2026cuaskill}. MMSkills differs by organizing skills around runtime state cards and multi-view evidence, and by using branch loading to align selected evidence with the live observation before the main agent acts.


\section{Conclusion and Limitations}
\label{sec:conclusion}

We introduced \textbf{MMSkills}, a framework that represents reusable skills for visual agents as multimodal procedural knowledge. By combining textual procedures, runtime state cards, multi-view keyframes, and branch-loaded use, MMSkills improve GUI and game-based visual agents across model families. The main limitations are dependence on source-trajectory coverage, possible errors from skill generation or visual grounding, and extra inference cost from branch loading. Extending MMSkills to broader embodied or safety-critical settings will require stronger verification and online skill repair.

\bibliographystyle{plainnat}
\bibliography{references}

\begin{thebibliography}{62}
\providecommand{\natexlab}[1]{#1}
\providecommand{\url}[1]{\texttt{#1}}
\expandafter\ifx\csname urlstyle\endcsname\relax
  \providecommand{\doi}[1]{doi: #1}\else
  \providecommand{\doi}{doi: \begingroup \urlstyle{rm}\Url}\fi

\bibitem[Agashe et~al.(2024)Agashe, Han, Gan, Yang, Li, and
  Wang]{agashe2024agents}
Saaket Agashe, Jiuzhou Han, Shuyu Gan, Jiachen Yang, Ang Li, and Xin~Eric Wang.
\newblock Agent {S}: An open agentic framework that uses computers like a
  human, 2024.
\newblock URL \url{https://arxiv.org/abs/2410.08164}.

\bibitem[Ahn et~al.(2022)Ahn, Brohan, Brown, Chebotar, Cortes, David, Finn, Fu,
  Gopalakrishnan, Hausman, Herzog, Ho, Hsu, Ibarz, Ichter, Irpan, Jang, Ruano,
  Jeffrey, Jesmonth, Joshi, Julian, Kalashnikov, Kuang, Lee, Levine, Lu, Luu,
  Parada, Pastor, Quiambao, Rao, Rettinghouse, Reyes, Sermanet, Sievers, Tan,
  Toshev, Vanhoucke, Xia, Xiao, Xu, Xu, Yan, and Zeng]{ichter2022saycan}
Michael Ahn, Anthony Brohan, Noah Brown, Yevgen Chebotar, Omar Cortes, Byron
  David, Chelsea Finn, Chuyuan Fu, Keerthana Gopalakrishnan, Karol Hausman,
  Alex Herzog, Daniel Ho, Jasmine Hsu, Julian Ibarz, Brian Ichter, Alex Irpan,
  Eric Jang, Rosario~Jauregui Ruano, Kyle Jeffrey, Sally Jesmonth, Nikhil~J
  Joshi, Ryan Julian, Dmitry Kalashnikov, Yuheng Kuang, Kuang-Huei Lee, Sergey
  Levine, Yao Lu, Linda Luu, Carolina Parada, Peter Pastor, Jornell Quiambao,
  Kanishka Rao, Jarek Rettinghouse, Diego Reyes, Pierre Sermanet, Nicolas
  Sievers, Clayton Tan, Alexander Toshev, Vincent Vanhoucke, Fei Xia, Ted Xiao,
  Peng Xu, Sichun Xu, Mengyuan Yan, and Andy Zeng.
\newblock Do as i can, not as i say: Grounding language in robotic affordances,
  2022.
\newblock URL \url{https://arxiv.org/abs/2204.01691}.

\bibitem[Alzubi et~al.(2026)Alzubi, Provenzano, Bingham, Chen, and
  Vu]{alzubi2026evoskill}
Salaheddin Alzubi, Noah Provenzano, Jaydon Bingham, Weiyuan Chen, and Tu~Vu.
\newblock Evoskill: Automated skill discovery for multi-agent systems, 2026.
\newblock URL \url{https://arxiv.org/abs/2603.02766}.

\bibitem[Bai et~al.(2025)Bai, Cai, Chen, Chen, Chen, Cheng, Deng, Ding, Gao,
  Ge, Ge, Guo, Huang, Huang, Huang, Hui, Jiang, Li, Li, Li, Li, Lin, Lin, Liu,
  Liu, Liu, Liu, Liu, Liu, Lu, Luo, Lv, Men, Meng, Ren, Ren, Song, Sun, Tang,
  Tu, Wan, Wang, Wang, Wang, Wang, Xie, Xu, Xu, Xu, Yang, Yang, Yang, Yang, Yu,
  Zhang, Zhang, Zhang, Zheng, Zhong, Zhou, Zhou, Zhou, Zhu, and
  Zhu]{bai2025qwen3vltechnicalreport}
Shuai Bai, Yuxuan Cai, Ruizhe Chen, Keqin Chen, Xionghui Chen, Zesen Cheng,
  Lianghao Deng, Wei Ding, Chang Gao, Chunjiang Ge, Wenbin Ge, Zhifang Guo,
  Qidong Huang, Jie Huang, Fei Huang, Binyuan Hui, Shutong Jiang, Zhaohai Li,
  Mingsheng Li, Mei Li, Kaixin Li, Zicheng Lin, Junyang Lin, Xuejing Liu,
  Jiawei Liu, Chenglong Liu, Yang Liu, Dayiheng Liu, Shixuan Liu, Dunjie Lu,
  Ruilin Luo, Chenxu Lv, Rui Men, Lingchen Meng, Xuancheng Ren, Xingzhang Ren,
  Sibo Song, Yuchong Sun, Jun Tang, Jianhong Tu, Jianqiang Wan, Peng Wang,
  Pengfei Wang, Qiuyue Wang, Yuxuan Wang, Tianbao Xie, Yiheng Xu, Haiyang Xu,
  Jin Xu, Zhibo Yang, Mingkun Yang, Jianxin Yang, An~Yang, Bowen Yu, Fei Zhang,
  Hang Zhang, Xi~Zhang, Bo~Zheng, Humen Zhong, Jingren Zhou, Fan Zhou, Jing
  Zhou, Yuanzhi Zhu, and Ke~Zhu.
\newblock Qwen3-{VL} technical report, 2025.
\newblock URL \url{https://arxiv.org/abs/2511.21631}.

\bibitem[Bai et~al.(2024)Bai, Lv, Zhang, Lyu, Tang, Huang, Du, Liu, Zeng, Hou,
  Dong, Tang, and Li]{bai2024longbenchbilingualmultitaskbenchmark}
Yushi Bai, Xin Lv, Jiajie Zhang, Hongchang Lyu, Jiankai Tang, Zhidian Huang,
  Zhengxiao Du, Xiao Liu, Aohan Zeng, Lei Hou, Yuxiao Dong, Jie Tang, and
  Juanzi Li.
\newblock Longbench: A bilingual, multitask benchmark for long context
  understanding, 2024.
\newblock URL \url{https://arxiv.org/abs/2308.14508}.

\bibitem[Chen et~al.(2026)Chen, Li, Solodko, Wang, Jiang, Cui, Hao, Ko, Abdali,
  Xu, Zheng, Fan, Cameron, Wagle, and Koishida]{chen2026cuaskill}
Tianyi Chen, Yinheng Li, Michael Solodko, Sen Wang, Nan Jiang, Tingyuan Cui,
  Junheng Hao, Jongwoo Ko, Sara Abdali, Leon Xu, Suzhen Zheng, Hao Fan,
  Pashmina Cameron, Justin Wagle, and Kazuhito Koishida.
\newblock {CUA}-skill: Develop skills for computer using agent, 2026.
\newblock URL \url{https://arxiv.org/abs/2601.21123}.

\bibitem[Cheng et~al.(2024)Cheng, Sun, Chu, Xu, Li, Zhang, and
  Wu]{cheng2024seeclick}
Kanzhi Cheng, Qiushi Sun, Yougang Chu, Fangzhi Xu, Yantao Li, Jianbing Zhang,
  and Zhiyong Wu.
\newblock Seeclick: Harnessing {GUI} grounding for advanced visual {GUI}
  agents.
\newblock In \emph{Proceedings of the 62nd Annual Meeting of the Association
  for Computational Linguistics}, pages 9313--9332. Association for
  Computational Linguistics, 2024.
\newblock \doi{10.18653/V1/2024.ACL-LONG.505}.
\newblock URL \url{https://doi.org/10.18653/v1/2024.acl-long.505}.

\bibitem[Deng et~al.(2023)Deng, Gu, Zheng, Chen, Stevens, Wang, Sun, and
  Su]{deng2023mind2web}
Xiang Deng, Yu~Gu, Boyuan Zheng, Shijie Chen, Samuel Stevens, Boshi Wang, Huan
  Sun, and Yu~Su.
\newblock Mind2web: Towards a generalist agent for the web, 2023.
\newblock URL \url{https://arxiv.org/abs/2306.06070}.

\bibitem[Gou et~al.(2025)Gou, Wang, Zheng, Xie, Chang, Shu, Sun, and
  Su]{gou2025navigatingdigitalworldhumans}
Boyu Gou, Ruohan Wang, Boyuan Zheng, Yanan Xie, Cheng Chang, Yiheng Shu, Huan
  Sun, and Yu~Su.
\newblock Navigating the digital world as humans do: Universal visual grounding
  for gui agents, 2025.
\newblock URL \url{https://arxiv.org/abs/2410.05243}.

\bibitem[He et~al.(2024)He, Yao, Ma, Yu, Dai, Zhang, Lan, and
  Yu]{he2024webvoyager}
Hongliang He, Wenlin Yao, Kaixin Ma, Wenhao Yu, Yong Dai, Hongming Zhang,
  Zhenzhong Lan, and Dong Yu.
\newblock Webvoyager: Building an end-to-end web agent with large multimodal
  models.
\newblock In \emph{Proceedings of the 62nd Annual Meeting of the Association
  for Computational Linguistics}, pages 6864--6890. Association for
  Computational Linguistics, 2024.
\newblock \doi{10.18653/V1/2024.ACL-LONG.371}.
\newblock URL \url{https://doi.org/10.18653/v1/2024.acl-long.371}.

\bibitem[Hong et~al.(2024)Hong, Wang, Lv, Xu, Yu, Ji, Wang, Wang, Zhang, Li,
  Xu, Dong, Ding, and Tang]{hong2024cogagentvisuallanguagemodel}
Wenyi Hong, Weihan Wang, Qingsong Lv, Jiazheng Xu, Wenmeng Yu, Junhui Ji, Yan
  Wang, Zihan Wang, Yuxuan Zhang, Juanzi Li, Bin Xu, Yuxiao Dong, Ming Ding,
  and Jie Tang.
\newblock Cogagent: A visual language model for gui agents, 2024.
\newblock URL \url{https://arxiv.org/abs/2312.08914}.

\bibitem[Hu et~al.(2025)Hu, Huo, Zhang, Yu, Xing, Stoica, Rosing, Jin, and
  Zhang]{hu2025lmgamebenchgoodllmsplaying}
Lanxiang Hu, Mingjia Huo, Yuxuan Zhang, Haoyang Yu, Eric~P. Xing, Ion Stoica,
  Tajana Rosing, Haojian Jin, and Hao Zhang.
\newblock lmgame-bench: How good are llms at playing games?, 2025.
\newblock URL \url{https://arxiv.org/abs/2505.15146}.

\bibitem[Ijspeert et~al.(2013)Ijspeert, Nakanishi, Hoffmann, Pastor, and
  Schaal]{ijspeert2013dmp}
Auke~Jan Ijspeert, Jun Nakanishi, Heiko Hoffmann, Peter Pastor, and Stefan
  Schaal.
\newblock Dynamical movement primitives: Learning attractor models for motor
  behaviors.
\newblock \emph{Neural Computation}, 25\penalty0 (2):\penalty0 328--373, 2013.
\newblock \doi{10.1162/NECO_a_00393}.
\newblock URL \url{https://doi.org/10.1162/NECO_a_00393}.

\bibitem[Jiang et~al.(2026)Jiang, Su, Qu, and
  Fung]{jiang2026xskillcontinuallearningexperience}
Guanyu Jiang, Zhaochen Su, Xiaoye Qu, and Yi~R. Fung.
\newblock Xskill: Continual learning from experience and skills in multimodal
  agents, 2026.
\newblock URL \url{https://arxiv.org/abs/2603.12056}.

\bibitem[Koh et~al.(2024)Koh, Lo, Jang, Duvvur, Lim, Huang, Neubig, Zhou,
  Salakhutdinov, and Fried]{koh2024visualwebarena}
Jing~Yu Koh, Robert Lo, Lawrence Jang, Vikram Duvvur, Ming~Chong Lim, Po-Yu
  Huang, Graham Neubig, Shuyan Zhou, Russ Salakhutdinov, and Daniel Fried.
\newblock Visualwebarena: Evaluating multimodal agents on realistic visual web
  tasks.
\newblock In \emph{Proceedings of the 62nd Annual Meeting of the Association
  for Computational Linguistics}, pages 881--905. Association for Computational
  Linguistics, 2024.
\newblock \doi{10.18653/V1/2024.ACL-LONG.50}.
\newblock URL \url{https://doi.org/10.18653/v1/2024.acl-long.50}.

\bibitem[Li et~al.(2025{\natexlab{a}})Li, Meng, Lin, Luo, Tian, Ma, Huang, and
  Chua]{li2025screenspotproguigroundingprofessional}
Kaixin Li, Ziyang Meng, Hongzhan Lin, Ziyang Luo, Yuchen Tian, Jing Ma, Zhiyong
  Huang, and Tat-Seng Chua.
\newblock Screenspot-pro: Gui grounding for professional high-resolution
  computer use, 2025{\natexlab{a}}.
\newblock URL \url{https://arxiv.org/abs/2504.07981}.

\bibitem[Li et~al.(2025{\natexlab{b}})Li, Xia, Du, Dai, Tang, Wang, Yu, and
  Zhang]{li2025rethinkmctsrefiningerroneousthoughts}
Qingyao Li, Wei Xia, Kounianhua Du, Xinyi Dai, Ruiming Tang, Yasheng Wang, Yong
  Yu, and Weinan Zhang.
\newblock Rethinkmcts: Refining erroneous thoughts in monte carlo tree search
  for code generation, 2025{\natexlab{b}}.
\newblock URL \url{https://arxiv.org/abs/2409.09584}.

\bibitem[Li et~al.(2026{\natexlab{a}})Li, Dai, Liu, Li, Wang, Tang, Yu, and
  Zhang]{li2026atgen}
Qingyao Li, Xinyi Dai, Weiwen Liu, Xiangyang Li, Yasheng Wang, Ruiming Tang,
  Yong Yu, and Weinan Zhang.
\newblock {ATG}en: Adversarial reinforcement learning for test case generation.
\newblock In \emph{The Fourteenth International Conference on Learning
  Representations}, 2026{\natexlab{a}}.
\newblock URL \url{https://openreview.net/forum?id=Sxj4o3qXtl}.

\bibitem[Li et~al.(2026{\natexlab{b}})Li, Chen, Liu, Zheng, Chen, He, Li, You,
  Shen, Sun, Wang, Li, Zeng, Wang, Zhao, Wang, Chaim, Di, Gao, He, He, Jing,
  Kong, Lan, Li, Li, Li, Lin, Liu, Liu, Lyu, Ma, Wang, Wang, Wang, Ye, Zhang,
  Xing, Xue, Dillmann, and chung Lee]{li2026skillsbenchbenchmarkingagentskills}
Xiangyi Li, Wenbo Chen, Yimin Liu, Shenghan Zheng, Xiaokun Chen, Yifeng He,
  Yubo Li, Bingran You, Haotian Shen, Jiankai Sun, Shuyi Wang, Binxu Li,
  Qunhong Zeng, Di~Wang, Xuandong Zhao, Yuanli Wang, Roey~Ben Chaim, Zonglin
  Di, Yipeng Gao, Junwei He, Yizhuo He, Liqiang Jing, Luyang Kong, Xin Lan,
  Jiachen Li, Songlin Li, Yijiang Li, Yueqian Lin, Xinyi Liu, Xuanqing Liu,
  Haoran Lyu, Ze~Ma, Bowei Wang, Runhui Wang, Tianyu Wang, Wengao Ye, Yue
  Zhang, Hanwen Xing, Yiqi Xue, Steven Dillmann, and Han chung Lee.
\newblock Skillsbench: Benchmarking how well agent skills work across diverse
  tasks, 2026{\natexlab{b}}.
\newblock URL \url{https://arxiv.org/abs/2602.12670}.

\bibitem[Liang et~al.(2023)Liang, Huang, Xia, Xu, Hausman, Ichter, Florence,
  and Zeng]{liang2023codepolicies}
Jacky Liang, Wenlong Huang, Fei Xia, Peng Xu, Karol Hausman, Brian Ichter, Pete
  Florence, and Andy Zeng.
\newblock Code as policies: Language model programs for embodied control.
\newblock In \emph{{IEEE} International Conference on Robotics and Automation,
  {ICRA} 2023}, pages 9493--9500. {IEEE}, 2023.
\newblock \doi{10.1109/ICRA48891.2023.10160591}.
\newblock URL \url{https://doi.org/10.1109/ICRA48891.2023.10160591}.

\bibitem[Liu et~al.(2023)Liu, Lin, Hewitt, Paranjape, Bevilacqua, Petroni, and
  Liang]{liu2023lostmiddlelanguagemodels}
Nelson~F. Liu, Kevin Lin, John Hewitt, Ashwin Paranjape, Michele Bevilacqua,
  Fabio Petroni, and Percy Liang.
\newblock Lost in the middle: How language models use long contexts, 2023.
\newblock URL \url{https://arxiv.org/abs/2307.03172}.

\bibitem[Liu et~al.(2024{\natexlab{a}})Liu, Zhang, Gu, Iong, Xu, Song, Zhang,
  Lai, Liu, Zhao, Sun, Yang, Yang, Qi, Yao, Sun, Cheng, Zheng, Yu, Zhang, Hong,
  Ding, Pan, Gu, Zeng, Du, Song, Su, Dong, and Tang]{liu2024visualagentbench}
Xiao Liu, Tianjie Zhang, Yu~Gu, Iat~Long Iong, Yifan Xu, Xixuan Song, Shudan
  Zhang, Hanyu Lai, Xinyi Liu, Hanlin Zhao, Jiadai Sun, Xinyue Yang, Yu~Yang,
  Zehan Qi, Shuntian Yao, Xueqiao Sun, Siyi Cheng, Qinkai Zheng, Hao Yu,
  Hanchen Zhang, Wenyi Hong, Ming Ding, Lihang Pan, Xiaotao Gu, Aohan Zeng,
  Zhengxiao Du, Chan~Hee Song, Yu~Su, Yuxiao Dong, and Jie Tang.
\newblock Visualagentbench: Towards large multimodal models as visual
  foundation agents, 2024{\natexlab{a}}.
\newblock URL \url{https://arxiv.org/abs/2408.06327}.

\bibitem[Liu et~al.(2024{\natexlab{b}})Liu, Zhang, Ren, Huang, Jin, Qin, Su,
  Xu, Yu, and Zhang]{liu2024alignrecaligningtrainingmultimodalrecommendations}
Yifan Liu, Kangning Zhang, Xiangyuan Ren, Yanhua Huang, Jiarui Jin, Yingjie
  Qin, Ruilong Su, Ruiwen Xu, Yong Yu, and Weinan Zhang.
\newblock Alignrec: Aligning and training in multimodal recommendations.
\newblock In \emph{Proceedings of the 33rd ACM International Conference on
  Information and Knowledge Management}, CIKM '24, pages 1503--1512, New York,
  NY, USA, 2024{\natexlab{b}}. Association for Computing Machinery.
\newblock ISBN 9798400704369.
\newblock \doi{10.1145/3627673.3679626}.
\newblock URL \url{https://doi.org/10.1145/3627673.3679626}.

\bibitem[Liu et~al.(2026)Liu, Ji, An, Jaakkola, Zhang, and
  Chang]{liu2026agenticskillsworkwild}
Yujian Liu, Jiabao Ji, Li~An, Tommi Jaakkola, Yang Zhang, and Shiyu Chang.
\newblock How well do agentic skills work in the wild: Benchmarking {LLM} skill
  usage in realistic settings, 2026.
\newblock URL \url{https://arxiv.org/abs/2604.04323}.

\bibitem[Lu et~al.(2024)Lu, Yang, Shen, and
  Awadallah]{lu2024omniparserpurevisionbased}
Yadong Lu, Jianwei Yang, Yelong Shen, and Ahmed Awadallah.
\newblock Omniparser for pure vision based gui agent, 2024.
\newblock URL \url{https://arxiv.org/abs/2408.00203}.

\bibitem[Ma et~al.(2026)Ma, Yang, Ji, Wang, Wang, Hu, Huang, and
  Chu]{ma2026skillclawletskillsevolve}
Ziyu Ma, Shidong Yang, Yuxiang Ji, Xucong Wang, Yong Wang, Yiming Hu, Tongwen
  Huang, and Xiangxiang Chu.
\newblock Skillclaw: Let skills evolve collectively with agentic evolver, 2026.
\newblock URL \url{https://arxiv.org/abs/2604.08377}.

\bibitem[Mayer(2009)]{mayer2009multimedia}
Richard~E. Mayer.
\newblock \emph{Multimedia Learning}.
\newblock Cambridge University Press, 2009.
\newblock \doi{10.1017/CBO9780511811678}.
\newblock URL \url{https://doi.org/10.1017/CBO9780511811678}.

\bibitem[Packer et~al.(2024)Packer, Wooders, Lin, Fang, Patil, Stoica, and
  Gonzalez]{packer2024memgptllmsoperatingsystems}
Charles Packer, Sarah Wooders, Kevin Lin, Vivian Fang, Shishir~G. Patil, Ion
  Stoica, and Joseph~E. Gonzalez.
\newblock Memgpt: Towards llms as operating systems, 2024.
\newblock URL \url{https://arxiv.org/abs/2310.08560}.

\bibitem[Park et~al.(2023)Park, O'Brien, Cai, Morris, Liang, and
  Bernstein]{park2023generativeagentsinteractivesimulacra}
Joon~Sung Park, Joseph~C. O'Brien, Carrie~J. Cai, Meredith~Ringel Morris, Percy
  Liang, and Michael~S. Bernstein.
\newblock Generative agents: Interactive simulacra of human behavior, 2023.
\newblock URL \url{https://arxiv.org/abs/2304.03442}.

\bibitem[Qin et~al.(2025)Qin, Ye, Fang, Wang, Liang, Tian, Zhang, Li, Li,
  Huang, Zhong, Li, Yang, Miao, Lin, Liu, Jiang, Ma, Li, Xiao, Cai, Li, Zheng,
  Jin, Li, Zhou, Wang, Chen, Li, Yang, Liu, Lin, Peng, Liu, and
  Shi]{qin2025uitars}
Yujia Qin, Yining Ye, Junjie Fang, Haoming Wang, Shihao Liang, Shizuo Tian,
  Junda Zhang, Jiahao Li, Yunxin Li, Shijue Huang, Wanjun Zhong, Kuanye Li,
  Jiale Yang, Yu~Miao, Woyu Lin, Longxiang Liu, Xu~Jiang, Qianli Ma, Jingyu Li,
  Xiaojun Xiao, Kai Cai, Chuang Li, Yaowei Zheng, Chaolin Jin, Chen Li, Xiao
  Zhou, Minchao Wang, Haoli Chen, Zhaojian Li, Haihua Yang, Haifeng Liu, Feng
  Lin, Tao Peng, Xin Liu, and Guang Shi.
\newblock {UI-TARS}: Pioneering automated {GUI} interaction with native agents,
  2025.
\newblock URL \url{https://arxiv.org/abs/2501.12326}.

\bibitem[Rawles et~al.(2023)Rawles, Li, Rodriguez, Riva, and
  Lillicrap]{rawles2023androidwild}
Christopher Rawles, Alice Li, Daniel Rodriguez, Oriana Riva, and Timothy
  Lillicrap.
\newblock Android in the wild: A large-scale dataset for android device
  control, 2023.
\newblock URL \url{https://arxiv.org/abs/2307.10088}.

\bibitem[Rawles et~al.(2025)Rawles, Clinckemaillie, Chang, Waltz, Lau, Fair,
  Li, Bishop, Li, Campbell-Ajala, Toyama, Berry, Tyamagundlu, Lillicrap, and
  Riva]{rawles2025androidworld}
Christopher Rawles, Sarah Clinckemaillie, Yifan Chang, Jonathan Waltz,
  Gabrielle Lau, Marybeth Fair, Alice Li, William Bishop, Wei Li, Folawiyo
  Campbell-Ajala, Daniel Toyama, Robert Berry, Divya Tyamagundlu, Timothy
  Lillicrap, and Oriana Riva.
\newblock Androidworld: A dynamic benchmarking environment for autonomous
  agents, 2025.
\newblock URL \url{https://arxiv.org/abs/2405.14573}.

\bibitem[Shao et~al.(2026{\natexlab{a}})Shao, Liu, Lu, and
  Zhang]{shao2026monoscalescalingmultiagentmonotonic}
Shuai Shao, Yixiang Liu, Bingwei Lu, and Weinan Zhang.
\newblock Monoscale: Scaling multi-agent system with monotonic improvement,
  2026{\natexlab{a}}.
\newblock URL \url{https://arxiv.org/abs/2601.23219}.

\bibitem[Shao et~al.(2026{\natexlab{b}})Shao, Ren, Qian, Wei, Guo, Yang, Song,
  Zhang, Zhang, Liu, and Shao]{shao2026agentmisevolveemergentrisks}
Shuai Shao, Qihan Ren, Chen Qian, Boyi Wei, Dadi Guo, Jingyi Yang, Xinhao Song,
  Linfeng Zhang, Weinan Zhang, Dongrui Liu, and Jing Shao.
\newblock Your agent may misevolve: Emergent risks in self-evolving llm agents,
  2026{\natexlab{b}}.
\newblock URL \url{https://arxiv.org/abs/2509.26354}.

\bibitem[Shinn et~al.(2023)Shinn, Cassano, Berman, Gopinath, Narasimhan, and
  Yao]{shinn2023reflexion}
Noah Shinn, Federico Cassano, Edward Berman, Ashwin Gopinath, Karthik
  Narasimhan, and Shunyu Yao.
\newblock Reflexion: language agents with verbal reinforcement learning.
\newblock In \emph{Advances in Neural Information Processing Systems 36}, 2023.
\newblock URL
  \url{http://papers.nips.cc/paper_files/paper/2023/hash/1b44b878bb782e6954cd888628510e90-Abstract-Conference.html}.

\bibitem[Sutton et~al.(1999)Sutton, Precup, and Singh]{sutton1999options}
Richard~S. Sutton, Doina Precup, and Satinder Singh.
\newblock Between {MDPs} and semi-{MDPs}: A framework for temporal abstraction
  in reinforcement learning.
\newblock \emph{Artificial Intelligence}, 112\penalty0 (1--2):\penalty0
  181--211, 1999.
\newblock \doi{10.1016/S0004-3702(99)00052-1}.
\newblock URL \url{https://doi.org/10.1016/S0004-3702(99)00052-1}.

\bibitem[Team et~al.(2026{\natexlab{a}})Team, Bai, Bai, Bao, Cai, Cao, Charles,
  Che, Chen, Chen, Chen, Chen, Chen, Chen, Chen, Chen, Chen, Chen, Chen, Chen,
  Chen, Chen, Chen, Chen, Chen, Chen, Chen, Chen, Chen, Cheng, Chu, Cui, Deng,
  Diao, Ding, Dong, Dong, Dong, Dong, Du, Du, Du, Du, Du, Fan, Fang, Feng,
  Feng, Fu, Fu, Gao, Gao, Ge, Geng, Gong, Gong, Gongque, Gu, Gu, Gu, Guan, Guo,
  Hao, He, He, He, Hong, Hu, Hu, Hu, Hu, Huang, Huang, Huang, Huang, Jiang,
  Jiang, Jin, Jing, Lai, Li, Li, Li, Li, Li, Li, Li, Li, Li, Li, Li, Li, Li,
  Li, Li, Li, Li, Li, Li, Li, Li, Li, Li, Liao, Lin, Lin, Lin, Lin, Liu, Liu,
  Liu, Liu, Liu, Liu, Liu, Liu, Liu, Liu, Liu, Liu, Liu, Liu, Liu, Liu, Liu,
  Liu, Lu, Lu, Lu, Luo, Luo, Luo, Ma, Ma, Mao, Mei, Men, Meng, Meng, Miao, Ni,
  Ouyang, Pan, Pang, Qian, Qin, Qin, Qiu, Qu, Shang, Shao, Shen, Shen, Shi,
  Shi, Shi, Song, Song, Song, Song, Su, Su, Su, Sui, Sun, Sun, Sun, Sung, Tai,
  Tang, Tang, Tang, Tang, Tao, Teng, Tian, Tian, Wang, Wang, Wang, Wang, Wang,
  Wang, Wang, Wang, Wang, Wang, Wang, Wang, Wang, Wang, Wang, Wang, Wang, Wang,
  Wang, Wang, Wang, Wang, Wang, Wang, Wang, Wang, Wang, Wang, Wang, Wang, Wang,
  Wang, Wang, Wang, Wang, Wang, Wang, Wang, Wei, Wei, Wen, Wen, Wu, Wu, Wu, Wu,
  Wu, Wu, Wu, Wu, Wu, Xiao, Xie, Xie, Xie, Xin, Xing, Xu, Xu, Xu, Xu, Xu, Xu,
  Xu, Xu, Xu, Xu, Xu, Xu, Xu, Xu, Xu, Yan, Yan, Yang, Yang, Yang, Yang, Yang,
  Yang, Yang, Yang, Yang, Yang, Yang, Yang, Yang, Yang, Yao, Ye, Ye, Ye, Yin,
  Yu, Yu, Yu, Yu, Yuan, Yuan, Yuan, Yue, Zeng, Zha, Zhan, Zhang, Zhang, Zhang,
  Zhang, Zhang, Zhang, Zhang, Zhang, Zhang, Zhang, Zhang, Zhang, Zhang, Zhang,
  Zhang, Zhang, Zhang, Zhang, Zhao, Zhao, Zhao, Zhao, Zhao, Zhao, Zhao, Zheng,
  Zheng, Zheng, Zheng, Zhong, Zhong, Zhong, Zhou, Zhou, Zhou, Zhou, Zhu, Zhu,
  Zhu, Zhu, Zhu, Zhuang, Zhuang, Zou, and Zu]{kimiteam2026kimik25visualagentic}
Kimi Team, Tongtong Bai, Yifan Bai, Yiping Bao, S.~H. Cai, Yuan Cao,
  Y.~Charles, H.~S. Che, Cheng Chen, Guanduo Chen, Huarong Chen, Jia Chen,
  Jiahao Chen, Jianlong Chen, Jun Chen, Kefan Chen, Liang Chen, Ruijue Chen,
  Xinhao Chen, Yanru Chen, Yanxu Chen, Yicun Chen, Yimin Chen, Yingjiang Chen,
  Yuankun Chen, Yujie Chen, Yutian Chen, Zhirong Chen, Ziwei Chen, Dazhi Cheng,
  Minghan Chu, Jialei Cui, Jiaqi Deng, Muxi Diao, Hao Ding, Mengfan Dong,
  Mengnan Dong, Yuxin Dong, Yuhao Dong, Angang Du, Chenzhuang Du, Dikang Du,
  Lingxiao Du, Yulun Du, Yu~Fan, Shengjun Fang, Qiulin Feng, Yichen Feng,
  Garimugai Fu, Kelin Fu, Hongcheng Gao, Tong Gao, Yuyao Ge, Shangyi Geng,
  Chengyang Gong, Xiaochen Gong, Zhuoma Gongque, Qizheng Gu, Xinran Gu, Yicheng
  Gu, Longyu Guan, Yuanying Guo, Xiaoru Hao, Weiran He, Wenyang He, Yunjia He,
  Chao Hong, Hao Hu, Jiaxi Hu, Yangyang Hu, Zhenxing Hu, Ke~Huang, Ruiyuan
  Huang, Weixiao Huang, Zhiqi Huang, Tao Jiang, Zhejun Jiang, Xinyi Jin,
  Yu~Jing, Guokun Lai, Aidi Li, C.~Li, Cheng Li, Fang Li, Guanghe Li, Guanyu
  Li, Haitao Li, Haoyang Li, Jia Li, Jingwei Li, Junxiong Li, Lincan Li, Mo~Li,
  Weihong Li, Wentao Li, Xinhang Li, Xinhao Li, Yang Li, Yanhao Li, Yiwei Li,
  Yuxiao Li, Zhaowei Li, Zheming Li, Weilong Liao, Jiawei Lin, Xiaohan Lin,
  Zhishan Lin, Zichao Lin, Cheng Liu, Chenyu Liu, Hongzhang Liu, Liang Liu,
  Shaowei Liu, Shudong Liu, Shuran Liu, Tianwei Liu, Tianyu Liu, Weizhou Liu,
  Xiangyan Liu, Yangyang Liu, Yanming Liu, Yibo Liu, Yuanxin Liu, Yue Liu,
  Zhengying Liu, Zhongnuo Liu, Enzhe Lu, Haoyu Lu, Zhiyuan Lu, Junyu Luo,
  Tongxu Luo, Yashuo Luo, Long Ma, Yingwei Ma, Shaoguang Mao, Yuan Mei, Xin
  Men, Fanqing Meng, Zhiyong Meng, Yibo Miao, Minqing Ni, Kun Ouyang, Siyuan
  Pan, Bo~Pang, Yuchao Qian, Ruoyu Qin, Zeyu Qin, Jiezhong Qiu, Bowen Qu, Zeyu
  Shang, Youbo Shao, Tianxiao Shen, Zhennan Shen, Juanfeng Shi, Lidong Shi,
  Shengyuan Shi, Feifan Song, Pengwei Song, Tianhui Song, Xiaoxi Song, Hongjin
  Su, Jianlin Su, Zhaochen Su, Lin Sui, Jinsong Sun, Junyao Sun, Tongyu Sun,
  Flood Sung, Yunpeng Tai, Chuning Tang, Heyi Tang, Xiaojuan Tang, Zhengyang
  Tang, Jiawen Tao, Shiyuan Teng, Chaoran Tian, Pengfei Tian, Ao~Wang, Bowen
  Wang, Chensi Wang, Chuang Wang, Congcong Wang, Dingkun Wang, Dinglu Wang,
  Dongliang Wang, Feng Wang, Hailong Wang, Haiming Wang, Hengzhi Wang, Huaqing
  Wang, Hui Wang, Jiahao Wang, Jinhong Wang, Jiuzheng Wang, Kaixin Wang, Linian
  Wang, Qibin Wang, Shengjie Wang, Shuyi Wang, Si~Wang, Wei Wang, Xiaochen
  Wang, Xinyuan Wang, Yao Wang, Yejie Wang, Yipu Wang, Yiqin Wang, Yucheng
  Wang, Yuzhi Wang, Zhaoji Wang, Zhaowei Wang, Zhengtao Wang, Zhexu Wang, Zihan
  Wang, Zizhe Wang, Chu Wei, Ming Wei, Chuan Wen, Zichen Wen, Chengjie Wu,
  Haoning Wu, Junyan Wu, Rucong Wu, Wenhao Wu, Yuefeng Wu, Yuhao Wu, Yuxin Wu,
  Zijian Wu, Chenjun Xiao, Jin Xie, Xiaotong Xie, Yuchong Xie, Yifei Xin, Bowei
  Xing, Boyu Xu, Jianfan Xu, Jing Xu, Jinjing Xu, L.~H. Xu, Lin Xu, Suting Xu,
  Weixin Xu, Xinbo Xu, Xinran Xu, Yangchuan Xu, Yichang Xu, Yuemeng Xu, Zelai
  Xu, Ziyao Xu, Junjie Yan, Yuzi Yan, Guangyao Yang, Hao Yang, Junwei Yang, Kai
  Yang, Ningyuan Yang, Ruihan Yang, Xiaofei Yang, Xinlong Yang, Ying Yang,
  Yi~Yang, Yi~Yang, Zhen Yang, Zhilin Yang, Zonghan Yang, Haotian Yao, Dan Ye,
  Wenjie Ye, Zhuorui Ye, Bohong Yin, Chengzhen Yu, Longhui Yu, Tao Yu,
  Tianxiang Yu, Enming Yuan, Mengjie Yuan, Xiaokun Yuan, Yang Yue, Weihao Zeng,
  Dunyuan Zha, Haobing Zhan, Dehao Zhang, Hao Zhang, Jin Zhang, Puqi Zhang,
  Qiao Zhang, Rui Zhang, Xiaobin Zhang, Y.~Zhang, Yadong Zhang, Yangkun Zhang,
  Yichi Zhang, Yizhi Zhang, Yongting Zhang, Yu~Zhang, Yushun Zhang, Yutao
  Zhang, Yutong Zhang, Zheng Zhang, Chenguang Zhao, Feifan Zhao, Jinxiang Zhao,
  Shuai Zhao, Xiangyu Zhao, Yikai Zhao, Zijia Zhao, Huabin Zheng, Ruihan Zheng,
  Shaojie Zheng, Tengyang Zheng, Junfeng Zhong, Longguang Zhong, Weiming Zhong,
  M.~Zhou, Runjie Zhou, Xinyu Zhou, Zaida Zhou, Jinguo Zhu, Liya Zhu, Xinhao
  Zhu, Yuxuan Zhu, Zhen Zhu, Jingze Zhuang, Weiyu Zhuang, Ying Zou, and Xinxing
  Zu.
\newblock Kimi k2.5: Visual agentic intelligence, 2026{\natexlab{a}}.
\newblock URL \url{https://arxiv.org/abs/2602.02276}.

\bibitem[Team et~al.(2026{\natexlab{b}})Team, Hong, Gu, Pan, Yang, Wang, Wang,
  Yue, Wang, Wang, Wang, Liu, Yu, Wang, Li, Duan, Yang, Lv, Liu, Pan, Ning, Ji,
  Wang, Chen, Xu, Zhu, Cheng, Qi, Gan, Wang, Yao, Dou, Zhou, Wang, Ge, Li, Hou,
  Xue, Wang, Qi, He, Zhang, Liu, Cen, Li, Wang, Yang, Liu, Lu, Xu, Wang, Zhao,
  Wang, Xue, Xu, Zhang, Liu, Liu, Zhao, Li, Tong, Zhang, Zhang, Yan, Zheng, Xu,
  Bao, lat Long~long, Xu, Fan, Qian, Chen, Lin, Sun, Zheng, Wang, Li, Liu, Xu,
  Yang, Zhang, Yin, Zhao, Wu, Shi, Lv, Jia, Li, Chen, Wang, Zhang, Liu, Xu, Li,
  Huang, Dong, and Tang]{vteam2026glm5vturbonativefoundationmodel}
V~Team, Wenyi Hong, Xiaotao Gu, Ziyang Pan, Zhen Yang, Yuting Wang, Yue Wang,
  Yuanchang Yue, Yu~Wang, Yanling Wang, Yan Wang, Xijun Liu, Wenmeng Yu, Weihan
  Wang, Wei Li, Shuaiqi Duan, Sheng Yang, Ruiliang Lv, Mingdao Liu, Lihang Pan,
  Ke~Ning, Junhui Ji, Jinjiang Wang, Jing Chen, Jiazheng Xu, Jiale Zhu, Jiale
  Cheng, Ji~Qi, Guobing Gan, Guo Wang, Cong Yao, Zijun Dou, Zihao Zhou, Zihan
  Wang, Zhiqi Ge, Zhijie Li, Zhenyu Hou, Zhao Xue, Zehui Wang, Zehan Qi, Zehai
  He, Yutao Zhang, Yusen Liu, Yukuo Cen, Yuchen Li, Yuan Wang, Yu~Yang, Yongbin
  Liu, Yijian Lu, Yifan Xu, Yanzi Wang, Yanxiao Zhao, Yanfeng Wang, Yadong Xue,
  Yabo Xu, Xinyu Zhang, Xinyu Liu, Xiao Liu, Wenyi Zhao, Wenkai Li, Tianyu
  Tong, Tianshu Zhang, Shudan Zhang, Shengdong Yan, Qinkai Zheng, Mingde Xu,
  Licheng Bao, lat Long~long, Jiaxing Xu, Jiaxin Fan, Jiawen Qian, Jiali Chen,
  Jiahui Lin, Jiadai Sun, Haozhi Zheng, Haoran Wang, Haochen Li, Hanyu Liu, Han
  Xu, Fan Yang, Dan Zhang, Da~Yin, Chuangxin Zhao, Chengcheng Wu, Boyan Shi,
  Bowen Lv, Bowei Jia, Bo~Li, Bin Chen, Baoxu Wang, Peng Zhang, Debing Liu, Bin
  Xu, Juanzi Li, Minlie Huang, Yuxiao Dong, and Jie Tang.
\newblock Glm-5v-turbo: Toward a native foundation model for multimodal agents,
  2026{\natexlab{b}}.
\newblock URL \url{https://arxiv.org/abs/2604.26752}.

\bibitem[Wang et~al.(2026{\natexlab{a}})Wang, Yu, Xie, Yao, Fang, Qiao, Cao,
  Zheng, Qi, Zhang, and Deng]{wang2026skillx}
Chenxi Wang, Zhuoyun Yu, Xin Xie, Wuguannan Yao, Runnan Fang, Shuofei Qiao,
  Kexin Cao, Guozhou Zheng, Xiang Qi, Peng Zhang, and Shumin Deng.
\newblock Skillx: Automatically constructing skill knowledge bases for agents,
  2026{\natexlab{a}}.
\newblock URL \url{https://arxiv.org/abs/2604.04804}.

\bibitem[Wang et~al.(2023{\natexlab{a}})Wang, Xie, Jiang, Mandlekar, Xiao, Zhu,
  Fan, and Anandkumar]{wang2023voyager}
Guanzhi Wang, Yuqi Xie, Yunfan Jiang, Ajay Mandlekar, Chaowei Xiao, Yuke Zhu,
  Linxi Fan, and Anima Anandkumar.
\newblock Voyager: An open-ended embodied agent with large language models,
  2023{\natexlab{a}}.
\newblock URL \url{https://arxiv.org/abs/2305.16291}.

\bibitem[Wang et~al.(2026{\natexlab{b}})Wang, Wang, and
  Xu]{wang2026skilltesterbenchmarkingutilitysecurity}
Leye Wang, Zixing Wang, and Anjie Xu.
\newblock Skilltester: Benchmarking utility and security of agent skills,
  2026{\natexlab{b}}.
\newblock URL \url{https://arxiv.org/abs/2603.28815}.

\bibitem[Wang et~al.(2026{\natexlab{c}})Wang, Jin, Fu, Yan, Wang, Hu, Wang, Li,
  Zhang, Yao, Jiao, Cheng, Lu, and
  Ge]{wang2026museagentmultimodalreasoningagent}
Shijian Wang, Jiarui Jin, Runhao Fu, Zexuan Yan, Xingjian Wang, Mengkang Hu,
  Eric Wang, Xiaoxi Li, Kangning Zhang, Li~Yao, Wenxiang Jiao, Xuelian Cheng,
  Yuan Lu, and Zongyuan Ge.
\newblock Museagent: A multimodal reasoning agent with stateful experiences,
  2026{\natexlab{c}}.
\newblock URL \url{https://arxiv.org/abs/2603.27813}.

\bibitem[Wang et~al.(2025{\natexlab{a}})Wang, Wang, Lu, Yang, Xie, Wang, Deng,
  Guo, Xu, Wu, Shen, Li, Li, Li, Chen, Zheng, Li, Lei, Cao, Fu, Shin, Shin, Hu,
  Wang, Chen, Ye, Zhang, Du, Hu, Chen, Zhou, Yao, Chen, Gu, Wang, Wang, Yang,
  Zhong, Sung, Charles, Yang, and
  Yu]{wang2025opencuaopenfoundationscomputeruse}
Xinyuan Wang, Bowen Wang, Dunjie Lu, Junlin Yang, Tianbao Xie, Junli Wang,
  Jiaqi Deng, Xiaole Guo, Yiheng Xu, Chen~Henry Wu, Zhennan Shen, Zhuokai Li,
  Ryan Li, Xiaochuan Li, Junda Chen, Boyuan Zheng, Peihang Li, Fangyu Lei,
  Ruisheng Cao, Yeqiao Fu, Dongchan Shin, Martin Shin, Jiarui Hu, Yuyan Wang,
  Jixuan Chen, Yuxiao Ye, Danyang Zhang, Dikang Du, Hao Hu, Huarong Chen, Zaida
  Zhou, Haotian Yao, Ziwei Chen, Qizheng Gu, Yipu Wang, Heng Wang, Diyi Yang,
  Victor Zhong, Flood Sung, Y.~Charles, Zhilin Yang, and Tao Yu.
\newblock Opencua: Open foundations for computer-use agents,
  2025{\natexlab{a}}.
\newblock URL \url{https://arxiv.org/abs/2508.09123}.

\bibitem[Wang et~al.(2025{\natexlab{b}})Wang, Wu, Xie, Ding, Yang, Li, Liu, Li,
  Dong, Chen, Wang, Zhao, Chen, Duan, Xie, Yang, Su, Yu, Huang, Liu, Zhang,
  Zhang, Yue, Su, Zhu, Shen, Dai, and
  Wang]{wang2025mmbenchguihierarchicalmultiplatformevaluation}
Xuehui Wang, Zhenyu Wu, JingJing Xie, Zichen Ding, Bowen Yang, Zehao Li,
  Zhaoyang Liu, Qingyun Li, Xuan Dong, Zhe Chen, Weiyun Wang, Xiangyu Zhao,
  Jixuan Chen, Haodong Duan, Tianbao Xie, Chenyu Yang, Shiqian Su, Yue Yu, Yuan
  Huang, Yiqian Liu, Xiao Zhang, Yanting Zhang, Xiangyu Yue, Weijie Su, Xizhou
  Zhu, Wei Shen, Jifeng Dai, and Wenhai Wang.
\newblock Mmbench-gui: Hierarchical multi-platform evaluation framework for gui
  agents, 2025{\natexlab{b}}.
\newblock URL \url{https://arxiv.org/abs/2507.19478}.

\bibitem[Wang et~al.(2023{\natexlab{b}})Wang, Cai, Liu, Jin, Hou, Zhang, Lin,
  He, Zheng, Yang, Ma, and Liang]{wang2023jarvis}
Zihao Wang, Shaofei Cai, Anji Liu, Yonggang Jin, Jinbing Hou, Bowei Zhang,
  Haowei Lin, Zhaofeng He, Zilong Zheng, Yaodong Yang, Xiaojian Ma, and Yitao
  Liang.
\newblock Jarvis-1: Open-world multi-task agents with memory-augmented
  multimodal language models, 2023{\natexlab{b}}.
\newblock URL \url{https://arxiv.org/abs/2311.05997}.

\bibitem[Wang et~al.(2024)Wang, Cai, Chen, Liu, Ma, and Liang]{wang2024deps}
Zihao Wang, Shaofei Cai, Guanzhou Chen, Anji Liu, Xiaojian Ma, and Yitao Liang.
\newblock Describe, explain, plan and select: Interactive planning with large
  language models enables open-world multi-task agents, 2024.
\newblock URL \url{https://arxiv.org/abs/2302.01560}.

\bibitem[Wu et~al.(2024)Wu, Wu, Xu, Wang, Sun, Jia, Cheng, Ding, Chen, Liang,
  and Qiao]{wu2024osatlas}
Zhiyong Wu, Zhenyu Wu, Fangzhi Xu, Yian Wang, Qiushi Sun, Chengyou Jia, Kanzhi
  Cheng, Zichen Ding, Liheng Chen, Paul~Pu Liang, and Yu~Qiao.
\newblock {OS-ATLAS}: A foundation action model for generalist {GUI} agents,
  2024.
\newblock URL \url{https://arxiv.org/abs/2410.23218}.

\bibitem[Xia et~al.(2026)Xia, Chen, Wang, Liu, Zeng, Wang, Han, Zhou, Zhao,
  Chen, Zheng, Xie, and Yao]{xia2026skillrlevolvingagentsrecursive}
Peng Xia, Jianwen Chen, Hanyang Wang, Jiaqi Liu, Kaide Zeng, Yu~Wang, Siwei
  Han, Yiyang Zhou, Xujiang Zhao, Haifeng Chen, Zeyu Zheng, Cihang Xie, and
  Huaxiu Yao.
\newblock Skillrl: Evolving agents via recursive skill-augmented reinforcement
  learning, 2026.
\newblock URL \url{https://arxiv.org/abs/2602.08234}.

\bibitem[Xie et~al.(2024)Xie, Zhang, Chen, Li, Zhao, Cao, Hua, Cheng, Shin,
  Lei, Liu, Xu, Zhou, Savarese, Xiong, Zhong, and Yu]{xie2024osworld}
Tianbao Xie, Danyang Zhang, Jixuan Chen, Xiaochuan Li, Siheng Zhao, Ruisheng
  Cao, Toh~Jing Hua, Zhoujun Cheng, Dongchan Shin, Fangyu Lei, Yitao Liu,
  Yiheng Xu, Shuyan Zhou, Silvio Savarese, Caiming Xiong, Victor Zhong, and Tao
  Yu.
\newblock Osworld: Benchmarking multimodal agents for open-ended tasks in real
  computer environments, 2024.
\newblock URL \url{https://arxiv.org/abs/2404.07972}.

\bibitem[Xie et~al.(2025)Xie, Li, Shao, Chen, Zhou, Li, Jiang, and
  Nie]{xie2025mirage}
Yuquan Xie, Zaijing Li, Rui Shao, Gongwei Chen, Kaiwen Zhou, Yinchuan Li,
  Dongmei Jiang, and Liqiang Nie.
\newblock Mirage-1: Augmenting and updating {GUI} agent with hierarchical
  multimodal skills, 2025.
\newblock URL \url{https://arxiv.org/abs/2506.10387}.

\bibitem[Xu and Yan(2026)]{xu2026agentskills}
Renjun Xu and Yang Yan.
\newblock Agent skills for large language models: Architecture, acquisition,
  security, and the path forward, 2026.
\newblock URL \url{https://arxiv.org/abs/2602.12430}.

\bibitem[Xu et~al.(2025)Xu, Yang, Chen, Wang, Chen, and
  Tang]{xu2025deskvisionlargescaledesktop}
Yibin Xu, Liang Yang, Hao Chen, Hua Wang, Zhi Chen, and Yaohua Tang.
\newblock Deskvision: Large scale desktop region captioning for advanced gui
  agents, 2025.
\newblock URL \url{https://arxiv.org/abs/2503.11170}.

\bibitem[Yang et~al.(2025{\natexlab{a}})Yang, Shao, Liu, and
  Shao]{yang2025riosworldbenchmarkingriskmultimodal}
Jingyi Yang, Shuai Shao, Dongrui Liu, and Jing Shao.
\newblock Riosworld: Benchmarking the risk of multimodal computer-use agents,
  2025{\natexlab{a}}.
\newblock URL \url{https://arxiv.org/abs/2506.00618}.

\bibitem[Yang et~al.(2025{\natexlab{b}})Yang, Ci, and Shou]{yang2025macosworld}
Pei Yang, Hai Ci, and Mike~Zheng Shou.
\newblock macosworld: A multilingual interactive benchmark for {GUI} agents,
  2025{\natexlab{b}}.
\newblock URL \url{https://arxiv.org/abs/2506.04135}.

\bibitem[Yang et~al.(2025{\natexlab{c}})Yang, Chai, Shao, Song, Qi, Rui, and
  Zhang]{yang2025agentnetdecentralizedevolutionarycoordination}
Yingxuan Yang, Huacan Chai, Shuai Shao, Yuanyi Song, Siyuan Qi, Renting Rui,
  and Weinan Zhang.
\newblock Agentnet: Decentralized evolutionary coordination for llm-based
  multi-agent systems, 2025{\natexlab{c}}.
\newblock URL \url{https://arxiv.org/abs/2504.00587}.

\bibitem[Yao et~al.(2023)Yao, Zhao, Yu, Du, Shafran, Narasimhan, and
  Cao]{yao2023react}
Shunyu Yao, Jeffrey Zhao, Dian Yu, Nan Du, Izhak Shafran, Karthik~R.
  Narasimhan, and Yuan Cao.
\newblock React: Synergizing reasoning and acting in language models.
\newblock In \emph{The Eleventh International Conference on Learning
  Representations, {ICLR} 2023}. OpenReview.net, 2023.
\newblock URL \url{https://openreview.net/forum?id=WE_vluYUL-X}.

\bibitem[Zhang et~al.(2023)Zhang, Yang, Liu, Han, Chen, Huang, Fu, and
  Yu]{zhang2023appagentmultimodalagentssmartphone}
Chi Zhang, Zhao Yang, Jiaxuan Liu, Yucheng Han, Xin Chen, Zebiao Huang, Bin Fu,
  and Gang Yu.
\newblock Appagent: Multimodal agents as smartphone users, 2023.
\newblock URL \url{https://arxiv.org/abs/2312.13771}.

\bibitem[Zhang et~al.(2024)Zhang, Qin, Jin, Liu, Su, Zhang, and
  Yu]{zhang2024dreamdualrepresentationlearningmodel}
Kangning Zhang, Yingjie Qin, Jiarui Jin, Yifan Liu, Ruilong Su, Weinan Zhang,
  and Yong Yu.
\newblock Dream: A dual representation learning model for multimodal
  recommendation, 2024.
\newblock URL \url{https://arxiv.org/abs/2404.11119}.

\bibitem[Zhang et~al.(2025)Zhang, Jiao, Du, Lu, Liu, Zhang, and
  Yu]{zhang2025looptoolclosingdatatrainingloop}
Kangning Zhang, Wenxiang Jiao, Kounianhua Du, Yuan Lu, Weiwen Liu, Weinan
  Zhang, and Yong Yu.
\newblock Looptool: Closing the data-training loop for robust llm tool calls,
  2025.
\newblock URL \url{https://arxiv.org/abs/2511.09148}.

\bibitem[Zheng et~al.(2024)Zheng, Gou, Kil, Sun, and
  Su]{zheng2024gpt4visiongeneralistwebagent}
Boyuan Zheng, Boyu Gou, Jihyung Kil, Huan Sun, and Yu~Su.
\newblock Gpt-4v(ision) is a generalist web agent, if grounded, 2024.
\newblock URL \url{https://arxiv.org/abs/2401.01614}.

\bibitem[Zheng et~al.(2025)Zheng, Fatemi, Jin, Wang, Gandhi, Song, Gu,
  Srinivasa, Liu, Neubig, and Su]{zheng2025skillweaver}
Boyuan Zheng, Michael~Y. Fatemi, Xiaolong Jin, Zora~Zhiruo Wang, Apurva Gandhi,
  Yueqi Song, Yu~Gu, Jayanth Srinivasa, Gaowen Liu, Graham Neubig, and Yu~Su.
\newblock Skillweaver: Web agents can self-improve by discovering and honing
  skills, 2025.
\newblock URL \url{https://arxiv.org/abs/2504.07079}.

\bibitem[Zhou et~al.(2024)Zhou, Xu, Zhu, Zhou, Lo, Sridhar, Cheng, Ou, Bisk,
  Fried, Alon, and Neubig]{zhou2024webarena}
Shuyan Zhou, Frank~F. Xu, Hao Zhu, Xuhui Zhou, Robert Lo, Abishek Sridhar,
  Xianyi Cheng, Tianyue Ou, Yonatan Bisk, Daniel Fried, Uri Alon, and Graham
  Neubig.
\newblock Webarena: A realistic web environment for building autonomous agents,
  2024.
\newblock URL \url{https://arxiv.org/abs/2307.13854}.

\end{thebibliography}

\newpage
\beginappendix

\section{Benchmark Statistics}
\label{app:benchmark-statistics}

We use four visual-agent benchmarks. \textbf{OSWorld} is the primary GUI benchmark and contains Ubuntu desktop tasks across browsers, office software, creative tools, media applications, system settings, code editors, email, and multi-application workflows \citep{xie2024osworld}. \textbf{macOSWorld} provides an auxiliary cross-operating-system GUI evaluation with file management, media, productivity, system/interface, and system-application tasks \citep{yang2025macosworld}. \textbf{VAB-Minecraft} is the Minecraft subset of VisualAgentBench and evaluates item-acquisition tasks that require visual grounding, inventory tracking, recipe reasoning, tool use, and handling failed actions \citep{liu2024visualagentbench}. \textbf{LMGame-Bench} evaluates game-playing agents through a unified interface \citep{hu2025lmgamebenchgoodllmsplaying}; we use Super Mario Bros because its recurring visual situations naturally align with reusable multimodal skills.

\begin{table}[h]
\centering
\caption{Test-case distributions for OSWorld and macOSWorld. OSWorld contains 360 test cases; macOSWorld contains 143 test cases. ``Share'' is the percentage of test cases in each domain within the corresponding benchmark.}
\label{tab:desktop-test-distribution}
\small
\setlength{\tabcolsep}{5pt}
\begin{tabular}{cccccc}
\toprule
Benchmark & Domain & Count & Share & Snapshot-en & Snapshot-apps \\
\midrule
OSWorld & Multi-app & 93 & 25.83 & -- & -- \\
OSWorld & LibreOffice Calc & 47 & 13.06 & -- & -- \\
OSWorld & LibreOffice Impress & 47 & 13.06 & -- & -- \\
OSWorld & Chrome & 45 & 12.50 & -- & -- \\
OSWorld & GIMP & 26 & 7.22 & -- & -- \\
OSWorld & OS & 24 & 6.67 & -- & -- \\
OSWorld & LibreOffice Writer & 23 & 6.39 & -- & -- \\
OSWorld & VS Code & 23 & 6.39 & -- & -- \\
OSWorld & VLC & 17 & 4.72 & -- & -- \\
OSWorld & Thunderbird & 15 & 4.17 & -- & -- \\
\midrule
macOSWorld & File management & 29 & 20.28 & 29 & 0 \\
macOSWorld & Media & 12 & 8.39 & 0 & 12 \\
macOSWorld & Productivity & 35 & 24.48 & 16 & 19 \\
macOSWorld & System and interface & 29 & 20.28 & 29 & 0 \\
macOSWorld & System apps & 38 & 26.57 & 38 & 0 \\
\bottomrule
\end{tabular}
\end{table}

\section{Skill Source Statistics}
\label{app:skill-source-statistics}

All MMSkills are extracted from non-test trajectories. For OSWorld and macOSWorld, we use the Ubuntu and macOS subsets of OpenCUA trajectories as GUI skill sources \citep{wang2025opencuaopenfoundationscomputeruse}. For macOS, the raw OpenCUA trajectories do not directly follow the five macOSWorld categories; we therefore perform additional clustering and relevance filtering before assigning trajectories to the analysis categories below.

\begin{table}[h]
\centering
\caption{OpenCUA trajectory statistics used for GUI skill extraction. ``Tasks'' counts source trajectories, ``Share'' is the within-platform percentage, and ``Clusters'' is the number of Phase-0 semantic trajectory clusters used for downstream skill planning.}
\label{tab:opencua-source-distribution}
\small
\setlength{\tabcolsep}{6pt}
\begin{tabular}{ccccc}
\toprule
Platform & Domain & Tasks & Share & Clusters \\
\midrule
Ubuntu & Chrome & 718 & 17.1 & 17 \\
Ubuntu & LibreOffice Impress & 605 & 14.4 & 11 \\
Ubuntu & VS Code & 605 & 14.4 & 4 \\
Ubuntu & OS & 497 & 11.8 & 2 \\
Ubuntu & GIMP & 492 & 11.7 & 14 \\
Ubuntu & LibreOffice Writer & 490 & 11.7 & 3 \\
Ubuntu & Thunderbird & 300 & 7.1 & 11 \\
Ubuntu & LibreOffice Calc & 298 & 7.1 & 3 \\
Ubuntu & VLC & 200 & 4.8 & 8 \\
\midrule
macOS & Productivity & 1,424 & 45.1 & 20 \\
macOS & System apps & 768 & 24.3 & 11 \\
macOS & File management & 341 & 10.8 & 9 \\
macOS & Media & 315 & 10.0 & 7 \\
macOS & System and interface & 309 & 9.8 & 12 \\
\bottomrule
\end{tabular}
\end{table}

\begin{table}[h]
\centering
\caption{OSWorld MMSkill package statistics. ``\#Skills'' counts unique skill packages, while ``Skills/Task'' reports the average number of skill matches assigned to evaluation tasks and therefore need not equal \#Skills/\#Tasks. Word statistics are median/mean over skill procedures. ``Full/Focus'' and ``Before/After'' report counts of those view types; ``Transition Cards'' counts state cards with at least one before/after transition view, with percentages over state cards. The Total/Avg row reports total counts and weighted averages; $\dagger$ marks a fitted value estimated from domain-level medians.}
\label{tab:osworld-skill-package-distribution}
\scriptsize
\setlength{\tabcolsep}{2.3pt}
\resizebox{\textwidth}{!}{%
\begin{tabular}{cccccccccccc}
\toprule
Domain & \#Tasks & \#Skills & Skills/Task & Words Med/Mean & \#Cards & Cards/Skill & \#Views & Views/Card & Full/Focus & Before/After & Transition Cards \\
\midrule
Chrome & 45 & 34 & 1.20 & 653 / 630.9 & 134 & 3.94 & 292 & 2.18 & 134/134 & 13/11 & 24 (17.9\%) \\
GIMP & 26 & 26 & 1.19 & 470 / 400.2 & 77 & 2.96 & 190 & 2.47 & 77/77 & 14/22 & 36 (46.8\%) \\
Calc & 47 & 26 & 1.36 & 278 / 278.1 & 79 & 3.04 & 184 & 2.33 & 79/79 & 7/19 & 26 (32.9\%) \\
Impress & 47 & 20 & 1.32 & 498 / 466.2 & 60 & 3.00 & 140 & 2.33 & 60/60 & 1/19 & 20 (33.3\%) \\
Writer & 23 & 23 & 1.13 & 264 / 289.2 & 71 & 3.09 & 144 & 2.03 & 71/71 & 1/1 & 2 (2.8\%) \\
Multi-apps & 93 & 20 & 1.19 & 574 / 502.0 & 82 & 4.10 & 164 & 2.00 & 82/82 & 0/0 & 0 (0.0\%) \\
OS & 24 & 37 & 1.21 & 544 / 539.8 & 139 & 3.76 & 283 & 2.04 & 139/139 & 5/0 & 5 (3.6\%) \\
Thunderbird & 15 & 25 & 1.20 & 508 / 542.5 & 87 & 3.48 & 192 & 2.21 & 87/84 & 6/15 & 21 (24.1\%) \\
VLC & 17 & 18 & 1.00 & 260 / 275.3 & 61 & 3.39 & 122 & 2.00 & 61/61 & 0/0 & 0 (0.0\%) \\
VS Code & 23 & 18 & 1.09 & 391 / 389.3 & 89 & 4.94 & 187 & 2.10 & 89/89 & 9/0 & 9 (10.1\%) \\
\midrule
\textbf{Total / Avg.} & \textbf{360} & \textbf{247} & \textbf{1.21} & \textbf{498.0$^\dagger$ / 447.8} & \textbf{879} & \textbf{3.56} & \textbf{1898} & \textbf{2.16} & \textbf{879/876} & \textbf{56/87} & \textbf{143 (16.3\%)} \\
\bottomrule
\end{tabular}
}
\end{table}

\begin{table}[h]
\centering
\caption{macOSWorld MMSkill package statistics. ``\#Skills'' counts unique skill packages, while ``Skills/Task'' reports the average number of skill matches assigned to evaluation tasks. Word statistics are median/mean over skill procedures. ``Full/Focus'' and ``Before/After'' report counts of those view types; ``Transition Cards'' counts state cards with at least one before/after transition view, with percentages over state cards.}
\label{tab:macosworld-skill-package-distribution}
\scriptsize
\setlength{\tabcolsep}{2.3pt}
\resizebox{\textwidth}{!}{%
\begin{tabular}{cccccccccccc}
\toprule
Domain & \#Tasks & \#Skills & Skills/Task & Words Med/Mean & \#Cards & Cards/Skill & \#Views & Views/Card & Full/Focus & Before/After & Transition Cards \\
\midrule
File management & 29 & 30 & 1.03 & 358 / 374.5 & 62 & 2.07 & 128 & 2.06 & 62/62 & 4/0 & 4 (6.5\%) \\
Media & 12 & 25 & 2.08 & 378 / 400.8 & 55 & 2.20 & 116 & 2.11 & 55/55 & 6/0 & 6 (10.9\%) \\
Productivity & 35 & 59 & 1.69 & 324 / 330.2 & 125 & 2.12 & 261 & 2.09 & 125/125 & 11/0 & 11 (8.8\%) \\
System/interface & 29 & 88 & 3.03 & 282 / 285.5 & 182 & 2.07 & 380 & 2.09 & 182/182 & 16/0 & 16 (8.8\%) \\
System apps & 38 & 46 & 1.21 & 347 / 352.0 & 98 & 2.13 & 212 & 2.16 & 98/98 & 6/10 & 16 (16.3\%) \\
\midrule
\textbf{Total / Avg.} & \textbf{143} & \textbf{248} & \textbf{1.73} & \textbf{324 / 330.9} & \textbf{522} & \textbf{2.10} & \textbf{1097} & \textbf{2.10} & \textbf{522/522} & \textbf{43/10} & \textbf{53 (10.2\%)} \\
\bottomrule
\end{tabular}
}
\end{table}

\begin{table}[h]
\centering
\caption{Game benchmark MMSkill package statistics. Word statistics are median/mean over skill procedures and plans. ``Full/Focus'' and ``Before/After'' report counts of those view types; ``Transition Cards'' counts state cards with at least one before/after transition view, with percentages over state cards. $\dagger$ marks a fitted value estimated from the available before/after view counts.}
\label{tab:game-skill-package-distribution}
\scriptsize
\setlength{\tabcolsep}{3pt}
\resizebox{\textwidth}{!}{%
\begin{tabular}{ccccccccccc}
\toprule
Benchmark & \#Skills & Skill Words Med/Mean & Plan Words Med/Mean & \#Cards & Cards/Skill & \#Views & Views/Card & Full/Focus & Before/After & Transition Cards \\
\midrule
VAB-Minecraft & 24 & 278.5 / 281.7 & 68.0 / 68.4 & 79 & 3.29 & 185 & 2.34 & 79/79 & 8/19 & 20 (25.3\%) \\
Super Mario Bros & 10 & 374.0 / 370.8 & 280.0 / 271.0 & 34 & 3.40 & 48$^\dagger$ & 1.41$^\dagger$ & 34/0 & 5/9 & 14 (41.2\%)$^\dagger$ \\
\bottomrule
\end{tabular}
}
\end{table}

For VAB-Minecraft, we use the official training set as the source for extracting multimodal skill packages. For Super Mario Bros from LMGame-Bench, MMSkills are extracted from multiple runs over four source cases. In both settings, the skill-source data are disjoint from the final evaluation cases.

\section{Experiment Details}
\label{app:experiment-details}

Across all evaluations, agents plan from visual environment observations rather than privileged state, using desktop screenshots for GUI tasks and game screenshots for game tasks. For OSWorld and macOSWorld, we run the full evaluations primarily on Amazon Web Services using the official benchmark images and task definitions. The agent interacts through the benchmark harness, and we use a maximum interaction budget of 20 steps for both GUI benchmarks. VAB-Minecraft and Super Mario Bros follow their official evaluation protocols.

For VAB-Minecraft, we use the official test set for evaluation. The training trajectories described in Appendix~\ref{app:skill-source-statistics} are used only to generate reusable procedures, state cards, and keyframes; no test episodes are used during skill construction.

For Super Mario Bros from LMGame-Bench, we split the available game cases into disjoint source and evaluation subsets. The source cases are described in Appendix~\ref{app:skill-source-statistics}, while a separate set of four held-out cases is used for final evaluation. This separation ensures that the generated skills capture reusable game situations rather than memorizing the measured episodes.

We evaluate both frontier and smaller multimodal models: Gemini 3.1 Pro, Gemini 3 Flash\footnote{\url{https://storage.googleapis.com/deepmind-media/Model-Cards/Gemini-3-Flash-Model-Card.pdf}}, Qwen3-VL-235B-A22B-Thinking \citep{bai2025qwen3vltechnicalreport}, GLM-5V-Turbo \citep{vteam2026glm5vturbonativefoundationmodel}, Kimi-K2.6 \citep{kimiteam2026kimik25visualagentic}, and Qwen3-VL-8B-Instruct \citep{bai2025qwen3vltechnicalreport}. For each base model, we compare \emph{no-skill}, \emph{text-only skill}, and \emph{MMSkills} conditions. Unless otherwise stated, skill conditions use branch loading: text-only skills use the same branch mechanism without state cards or images, while MMSkills inspect selected state cards and multi-view keyframes before returning structured guidance to the main agent. Direct text-skill loading and direct multimodal loading are evaluated only as ablation variants.

\section{Branch-Loaded Runtime Algorithm}
\label{app:branch-loaded-algorithm}

Algorithm~\ref{alg:branch-loaded-agent} summarizes the branch-loaded runtime loop. Candidate skills are selected before task execution, while branch calls occur only when the main agent decides to consult a specific skill. The main trajectory receives the structured guidance $G_t$ rather than the full multimodal skill package.

\begin{algorithm}[t]
\caption{Branch-loaded MMSkill Agent}
\label{alg:branch-loaded-agent}
\begin{algorithmic}[1]
\Require Skill library $\mathcal{M}$, task instruction $I$, visual environment $\mfield{Env}$
\State Initialize history $H_0 \leftarrow \emptyset$
\State Pre-recall candidate skills $\mathcal{C}_I \leftarrow \text{PreRecall}(I,\mathcal{M})$
\For{$t=1,2,\ldots$}
    \State Observe current visual observation $O_t$ from $\mfield{Env}$
    \State Main agent chooses either action $A_t$ or skill request $M_t\in\mathcal{C}_I$
    \If{the main agent chooses action $A_t$}
        \State Execute $A_t$ in $\mfield{Env}$ and update $H_t$
    \Else
        \State Unpack $M_t=(D_t,P_t,S_t,K_t)$
        \State Stage 1: $(J_t,R_t)\leftarrow\text{SelectViews}(O_t,H_{t-1},P_t,S_t)$
        \State Load $V_t\leftarrow\{K_j^v:j\in J_t,\ v\in R_{t,j}\}$
        \State Stage 2: $G_t\leftarrow\text{PlanBranch}(O_t,H_{t-1},P_t,\{S_j:j\in J_t\},V_t)$
        \State Choose grounded action $A_t\leftarrow\pi_{\text{main}}(O_t,H_{t-1},G_t)$
        \State Execute $A_t$ in $\mfield{Env}$ and update $H_t$
    \EndIf
    \If{the task is verified complete}
        \State \Return success
    \EndIf
\EndFor
\end{algorithmic}
\end{algorithm}

\begin{figure}[h]
    \centering
    \includegraphics[width=\textwidth]{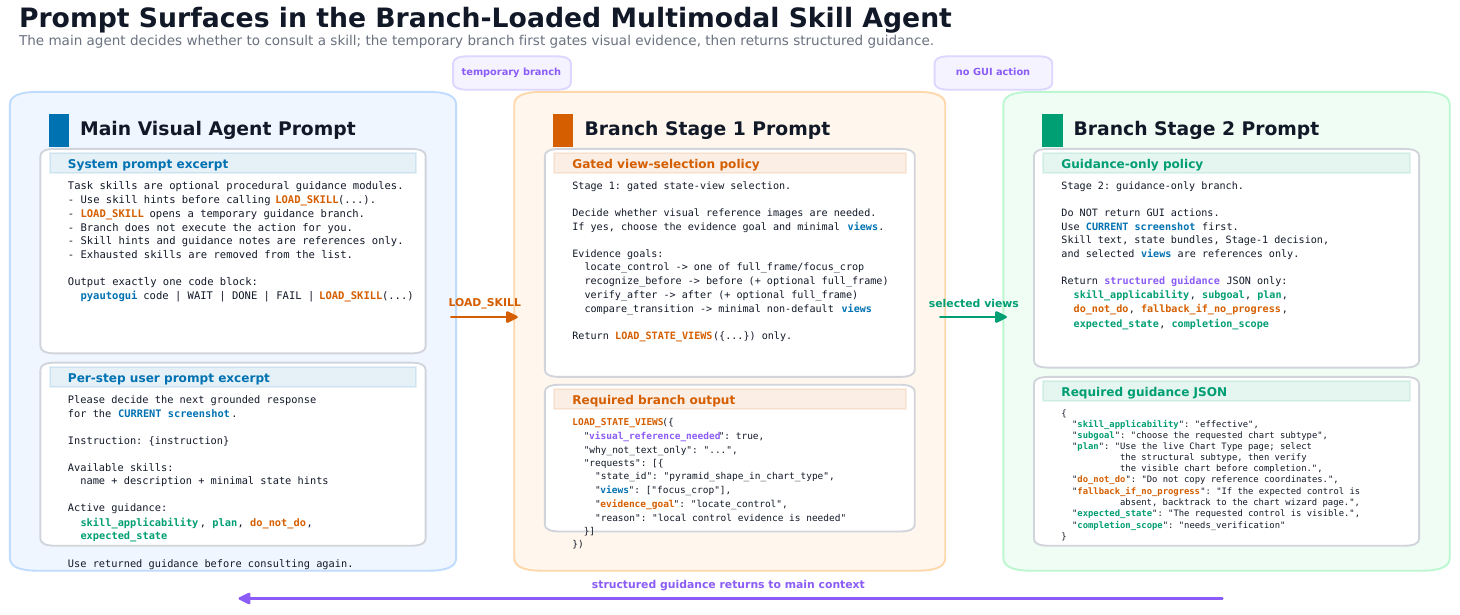}
    \caption{Prompt surfaces used by the branch-loaded multimodal skill agent. The main agent prompt decides whether to act directly or consult a skill branch, Stage 1 selects the relevant state cards and keyframe views, and Stage 2 returns compact structured guidance to the main agent.}
    \label{fig:appendix-agent-prompt-examples}
\end{figure}

\makeatletter
\ifdefined\promptbox\else
\newenvironment{promptbox}[2][]{%
    \par\medskip\noindent\begin{minipage}{\textwidth}
    \hrule\medskip\noindent\textbf{#2}\par\smallskip\small
}{%
    \medskip\hrule\end{minipage}\par\medskip
}
\fi
\makeatother

\section{MMSkillAgent Prompt Templates}
\label{app:mmskillagent-prompts}

This section reports the prompt templates used by the branch-loaded MMSkillAgent. Dynamic fields are shown as placeholders such as \texttt{\{instruction\}}, \texttt{\{available\_skills\}}, and \texttt{\{previous\_steps\}}. The implementation instantiates these templates with the current screenshot, recent trajectory, execution feedback, candidate skills, state-card summaries, and selected keyframe views. The Stage-2 JSON contains a few implementation-facing fields beyond Eq.~\ref{eq:branch-summary}; they are collapsed into $G_t$ in the method description.

\begin{promptbox}{Main-Agent Skill-Calling System Prompt}
\textbf{Role.} Follow the user instruction to perform desktop computer tasks. You control the computer using Python code with \texttt{pyautogui}. At each step, you receive the current screenshot and recent visible trajectory history. Use the current screenshot to decide the next action; do not assume previous clicks succeeded.

\medskip
\textbf{Skill consultation policy.}
\begin{itemize}[leftmargin=*, itemsep=0.2em]
    \item Task skills are optional procedural planners only.
    \item The final user message includes each non-exhausted skill's short description and minimal runtime state hints. Use these hints to judge whether a skill is genuinely relevant before calling \texttt{LOAD\_SKILL(...)}.
    \item Call \texttt{LOAD\_SKILL("<exact\_skill\_name>")} only when the current screenshot, recent steps, and skill hints suggest that extra procedural guidance is useful.
    \item \texttt{LOAD\_SKILL(...)} opens a temporary planner branch for extra skill-guided reasoning; it does not execute the action.
    \item Skill hints and planner notes are references only, never coordinate templates.
    \item Each skill may be consulted at most \texttt{\{consult\_limit\}} times in one trajectory. Exhausted skills are removed from the available-skill list and must not be called again.
\end{itemize}

\textbf{Available skills.} \texttt{\{available\_skills\}} lists non-exhausted candidate skills for the task.

\medskip
\textbf{Action rules.}
\begin{itemize}[leftmargin=*, itemsep=0.2em]
    \item Use \texttt{pyautogui} only for GUI actions. Do not use \texttt{pyautogui.locateCenterOnScreen} or \texttt{pyautogui.screenshot()}.
    \item Each response must be self-contained and must not rely on variables from previous steps.
    \item If a click does not work, revise the target from the new screenshot instead of repeating the same guess.
    \item Prefer short, direct, grounded actions over long speculative scripts; avoid repetitive unproductive loops.
    \item Before outputting \texttt{DONE}, verify that the full user instruction has been completed, not only a local subgoal.
\end{itemize}

\textbf{Output interface.} Return exactly one code block containing one of: Python code using \texttt{pyautogui}, \texttt{WAIT}, \texttt{DONE}, \texttt{FAIL}, or \texttt{LOAD\_SKILL("<exact\_skill\_name>")}. Do not mix Python code with a skill call, do not load more than one skill, and do not return prose outside the code block. If returning Python, include concise \texttt{\#} comments. Use \texttt{WAIT} only for loading UI, \texttt{DONE} only after full verification, and \texttt{FAIL} only when the task is truly impossible. Canonical outputs include \texttt{LOAD\_SKILL("Example\_Skill\_Name")} and a single grounded action such as \texttt{pyautogui.click(120, 54)}.

\medskip
\textbf{Coordinate and task context.} Use the declared screen resolution for all \texttt{pyautogui} coordinates. The computer password is available as \texttt{\{client\_password\}} when needed. The task is \texttt{\{instruction\}}.
\end{promptbox}

\begin{promptbox}{Main-Agent Per-Step User Instruction}
\textbf{Decision request.} Decide the next grounded response for the current screenshot. Return either the next GUI action or \texttt{LOAD\_SKILL(...)} when extra procedural guidance is useful.

\medskip
\textbf{Per-step context.}
\begin{itemize}[leftmargin=*, itemsep=0.2em]
    \item \textbf{Instruction:} \texttt{\{instruction\}}
    \item \textbf{Available non-exhausted skills:} \texttt{\{skills\_with\_state\_previews\}}, including each skill name, short description, and minimal when-to-use state hints.
    \item \textbf{Active planner memo:} \texttt{\{active\_memo\}}
    \item \textbf{Planner notes returned in this step:} \texttt{\{current\_step\_planner\_summaries\}}
    \item \textbf{Previous steps:} \texttt{\{previous\_steps\}}, including full model responses and action comments.
    \item \textbf{Execution feedback:} optional feedback for the current step and optional loop-warning diagnostics.
    \item \textbf{Screen resolution:} \texttt{\{screen\_resolution\_prompt\}}.
\end{itemize}

\textbf{Grounding rules.}
\begin{itemize}[leftmargin=*, itemsep=0.2em]
    \item Ground every action in the current screenshot.
    \item Planner notes are fallible references; re-decide the real action from the current screenshot, recent history, and execution feedback.
    \item Treat state hints, selected reference views, and planner notes as references only, never coordinate templates.
    \item If no listed skill is clearly useful, act directly from the current screenshot.
    \item If planner notes already exist for this step, use them before consulting another branch.
    \item If recent actions repeat without progress, change strategy.
    \item Before \texttt{DONE}, verify the full instruction; if returning Python, include concise comments.
\end{itemize}
\end{promptbox}

\begin{promptbox}{Branch Stage 1 Prompt: Gated State-View Selection}
\textbf{Branch reference package.} The branch receives the requested call \texttt{LOAD\_SKILL("\{skill\_name\}")}, the selected skill text, runtime state bundles, and compact state-card manifests. These materials are supplemental procedural references only. Stage 1 must decide whether visual reference images are needed at all and, if so, which state IDs and view types should be loaded. The main agent, not the branch, will choose the concrete GUI action.

\medskip
\textbf{Role.} You are inside Stage 1 of a temporary state-view selection branch for a single desktop step. Decide whether visual reference images are needed before planner reasoning and which evidence goal they should serve.

\medskip
\textbf{View semantics.}
\begin{itemize}[leftmargin=*, itemsep=0.2em]
    \item \texttt{full\_frame}: global placement and window context.
    \item \texttt{focus\_crop}: detailed control localization.
    \item \texttt{before}: pre-change state, useful for recognizing whether the UI is still before a change and for avoiding repeated toggles.
    \item \texttt{after}: target completion state, useful for verifying the result after save, enable, format, or apply operations.
\end{itemize}

\textbf{Evidence goals.}
\begin{itemize}[leftmargin=*, itemsep=0.2em]
    \item \texttt{locate\_control}: request exactly one of \texttt{full\_frame} or \texttt{focus\_crop}.
    \item \texttt{recognize\_before}: request \texttt{before}, optionally with \texttt{full\_frame}.
    \item \texttt{verify\_after}: request \texttt{after}, optionally with \texttt{full\_frame}.
    \item \texttt{compare\_transition}: request minimal transition evidence; avoid defaulting to the \texttt{full\_frame}+\texttt{focus\_crop} pair and prefer \texttt{before}/\texttt{after} when useful.
\end{itemize}

\textbf{Visual gating policy.} First decide \texttt{visual\_reference\_needed}. If the useful help is a generic shortcut, formula, file operation, stable menu path, or textual procedure, default to \texttt{false}. Load images only for state transitions, visual result verification, or complex UI-state recognition where text alone is likely insufficient. Keep the request minimal: at most \texttt{\{max\_states\}} states and \texttt{\{max\_views\}} total views.

\medskip
\textbf{Input fields.} Stage 1 receives \texttt{\{instruction\}}, \texttt{\{previous\_steps\}}, environment feedback from the previous step, loop warnings if present, the screen-resolution prompt, and the current screenshot.

\medskip
\textbf{Output interface.} Return exactly one code block containing one \texttt{LOAD\_STATE\_VIEWS(...)} call. Its JSON payload contains:
\begin{itemize}[leftmargin=*, itemsep=0.2em]
    \item \texttt{"visual\_reference\_needed"}: true or false;
    \item \texttt{"why\_not\_text\_only"}: why text-only is insufficient, or why no images are needed;
    \item \texttt{"requests"}: a list of objects, each with exact \texttt{"state\_id"}, exact \texttt{"views"}, \texttt{"evidence\_goal"}, and \texttt{"reason"}.
\end{itemize}
When \texttt{"visual\_reference\_needed"} is false, \texttt{"requests"} must be empty. Do not return Python code, planner JSON, \texttt{WAIT}, \texttt{DONE}, \texttt{FAIL}, \texttt{LOAD\_SKILL}, or prose outside the code block.

\medskip
\textbf{Canonical examples.} A transition request sets \texttt{"visual\_reference\_needed": true} and requests a state with \texttt{"views": ["before", "after"]} under \texttt{"evidence\_goal": "compare\_transition"}. A text-only branch sets \texttt{"visual\_reference\_needed": false}, gives a brief reason in \texttt{"why\_not\_text\_only"}, and returns \texttt{"requests": []}.
\end{promptbox}

\begin{promptbox}{Branch Stage 2 Prompt: Planner JSON}
\textbf{Selected evidence package.} Stage 2 receives the Stage-1 selection record, including the evidence goal, selected states, requested view types, reasons, when-to-use conditions, verification cues, and any loaded keyframe views. Loaded views are supplemental references only and are never coordinate templates.

\medskip
\textbf{Role.} You are inside Stage 2 of a temporary planner-only skill consultation branch for a single desktop step. Do not return a GUI action. Return a structured planner summary for the current state.

\medskip
\textbf{Branch rules.}
\begin{itemize}[leftmargin=*, itemsep=0.2em]
    \item Do not return Python code, \texttt{WAIT}, \texttt{DONE}, \texttt{FAIL}, \texttt{LOAD\_SKILL}, \texttt{LOAD\_SKILL\_IMAGE}, or \texttt{LOAD\_STATE\_VIEWS}. Do not request another skill in this branch.
    \item Use the current screenshot first. Skill text, runtime state bundles, Stage-1 decisions, and loaded reference views are supplemental only.
    \item If Stage 1 chose no visual references, respect that decision and avoid inventing image-based assumptions.
    \item If the skill is ineffective for the current state, say so clearly and avoid forcing the plan toward it.
    \item Treat reference views as state references, never as coordinate templates.
\end{itemize}

\textbf{Planning requirements.}
\begin{itemize}[leftmargin=*, itemsep=0.2em]
    \item \texttt{subgoal}: next immediate local milestone under the live UI.
    \item \texttt{plan}: longer-range route grounded in the current screenshot, including the relevant UI surface, the next 2--4 actions/checks/transitions, and the cue that means advance versus re-plan.
    \item \texttt{do\_not\_do}: the likely wrong path or skill-induced mistake to avoid.
    \item \texttt{fallback\_if\_no\_progress}: a concrete alternate route if the skill-guided path stalls.
    \item \texttt{expected\_state}: visible screenshot cues the main agent should aim to reveal next.
    \item \texttt{completion\_scope}: whether the branch only advances a local step, still needs verification, or may be complete after verification.
\end{itemize}

\textbf{Per-step input fields.} Stage 2 receives \texttt{\{instruction\}}, \texttt{\{stage1\_decision\}}, \texttt{\{selected\_state\_views\}}, \texttt{\{previous\_steps\}}, environment feedback, optional loop warnings, the screen-resolution prompt, and the live screenshot, which is more authoritative than any skill reference view.

\medskip
\textbf{Output interface.} Return exactly one code block containing one JSON object with keys:
\texttt{"skill\_applicability"}, \texttt{"subgoal"}, \texttt{"plan"}, \texttt{"do\_not\_do"}, \texttt{"fallback\_if\_no\_progress"}, \texttt{"expected\_state"}, and \texttt{"completion\_scope"}. The values of \texttt{"skill\_applicability"} are \texttt{"effective"}, \texttt{"ineffective"}, or \texttt{"uncertain"}; the values of \texttt{"completion\_scope"} are \texttt{"local\_only"}, \texttt{"needs\_verification"}, or \texttt{"maybe\_complete"}. Do not return prose outside the code block.

\medskip
\textbf{Canonical example shape.} A valid planner object may mark the skill as \texttt{"effective"}, set a local \texttt{"subgoal"} such as opening the visible settings surface, give a grounded multi-step \texttt{"plan"}, block a likely repeated or irrelevant click through \texttt{"do\_not\_do"}, provide a concrete fallback route, and describe the next visible \texttt{"expected\_state"} with \texttt{"completion\_scope": "local\_only"}.
\end{promptbox}

\section{Additional Behavioral Shift Analysis}
\label{app:additional-behavior-analysis}

\begin{figure}[h]
    \centering
    \includegraphics[width=\textwidth]{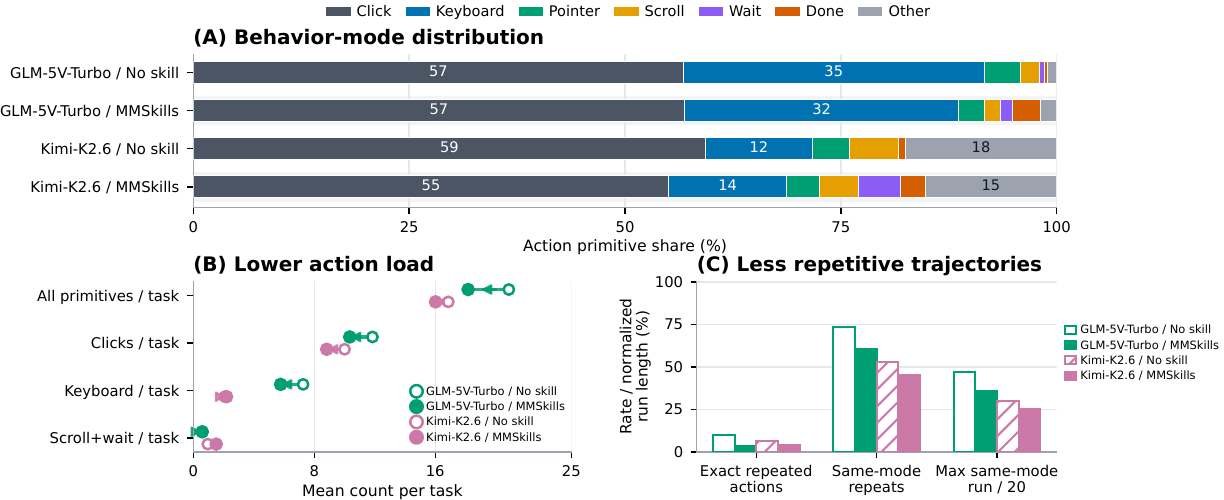}
    \caption{Behavioral shifts induced by MMSkills on OSWorld for GLM-5V and Kimi-K2.6. The panels follow the same metrics as Figure~\ref{fig:behavior-shift}: action primitive distribution, low-level primitives per task, and repetitive behavior statistics.}
    \label{fig:appendix-behavior-shift-glm-kimi}
\end{figure}

Figure~\ref{fig:appendix-behavior-shift-glm-kimi} complements Figure~\ref{fig:behavior-shift} with the same OSWorld behavioral analysis for GLM-5V and Kimi-K2.6.

\section{Interaction Case Studies}
\label{app:interaction-case-studies}

\begin{figure}[p]
    \centering
    \includegraphics[width=\textwidth,height=0.88\textheight,keepaspectratio]{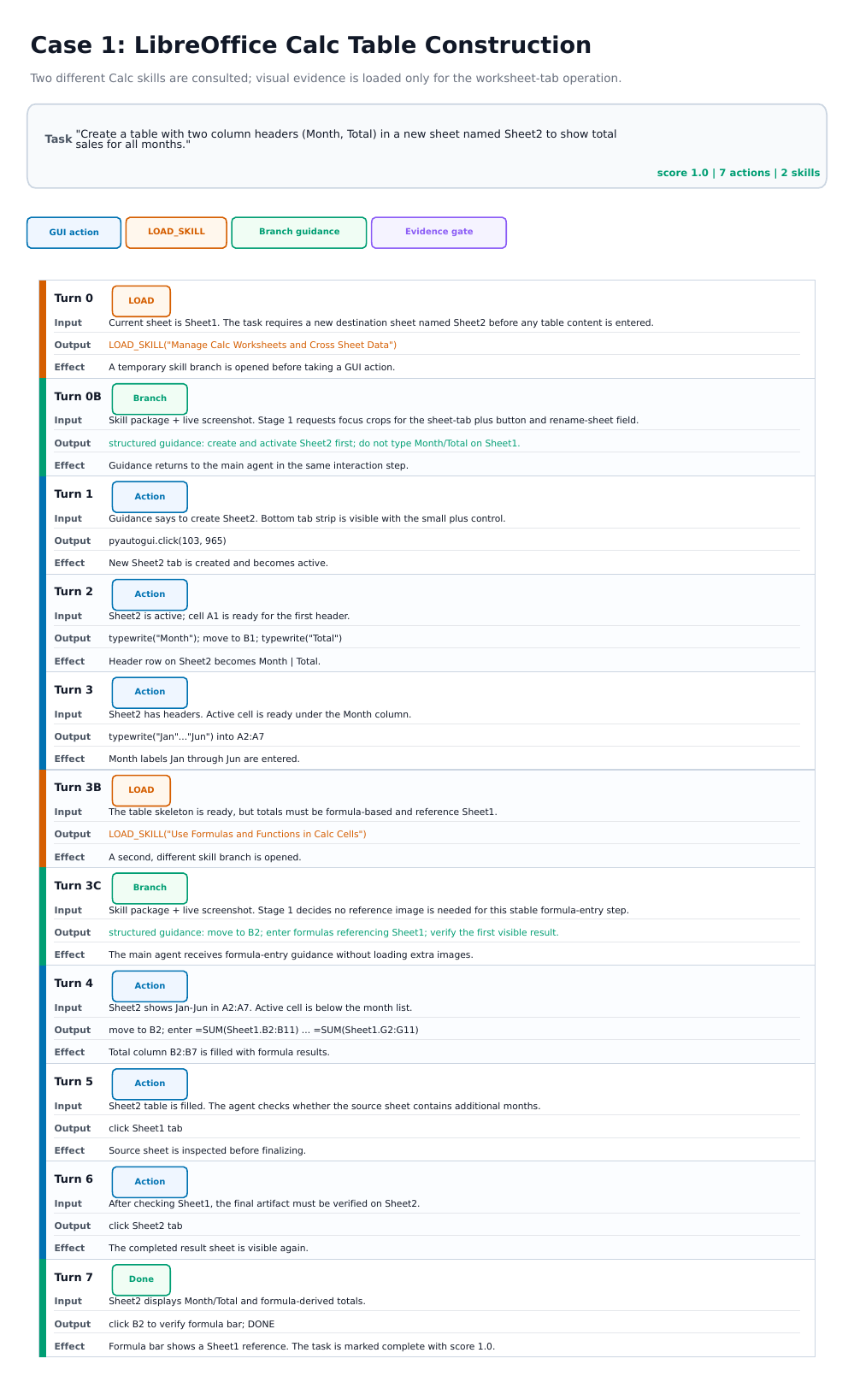}
    \caption{Representative interaction case with branch-loaded MMSkills: LibreOffice Calc table construction. Colored turn labels distinguish direct GUI actions, skill loading, branch guidance, evidence-gated reasoning, and final completion.}
    \label{fig:appendix-interaction-cases}
\end{figure}

\begin{figure}[p]
    \centering
    \includegraphics[width=\textwidth,height=0.88\textheight,keepaspectratio]{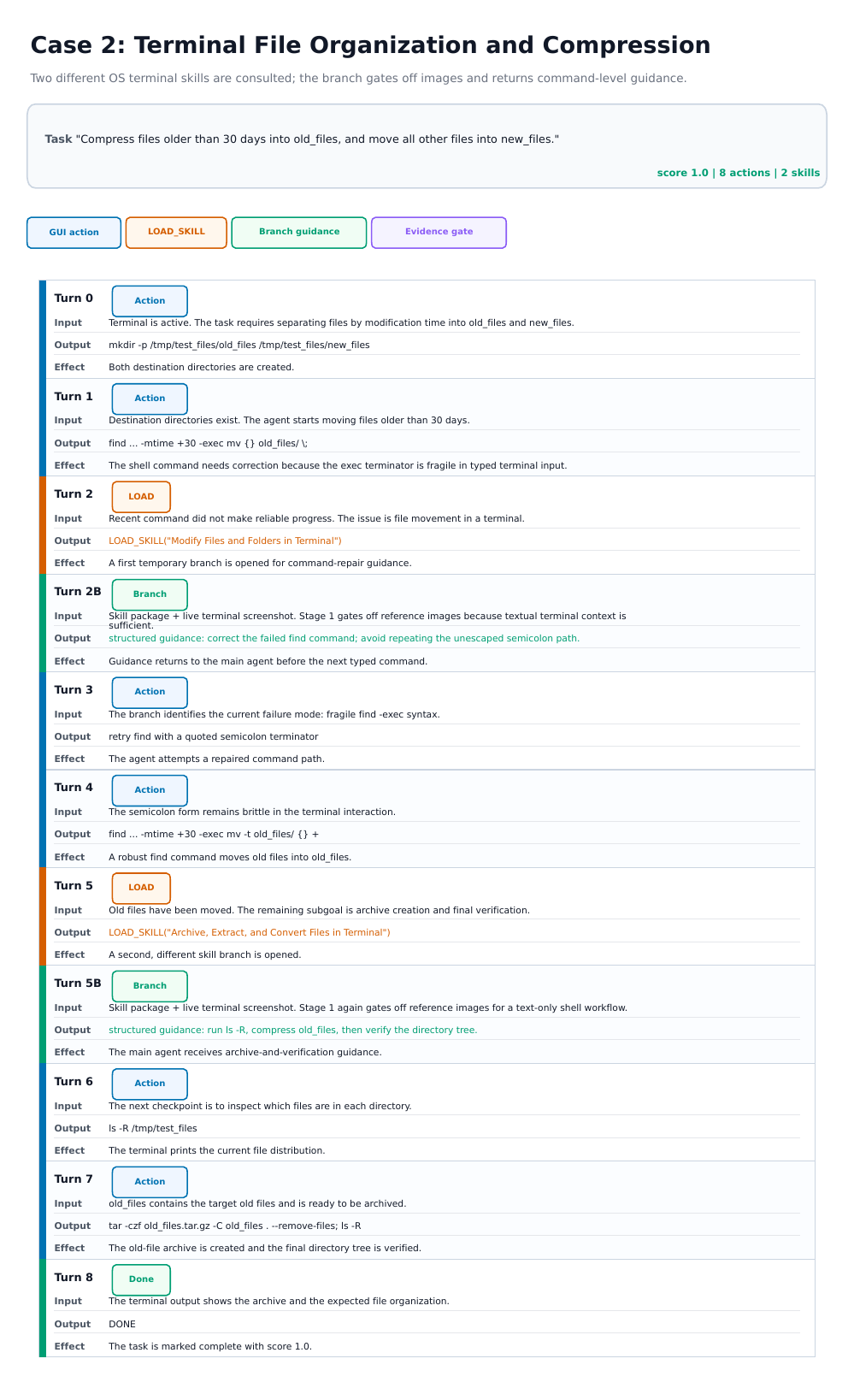}
    \caption{Representative interaction case with branch-loaded MMSkills: terminal file organization and compression. Colored turn labels distinguish direct GUI actions, skill loading, branch guidance, evidence-gated reasoning, and final completion.}
    \label{fig:appendix-interaction-case-terminal}
\end{figure}

Figures~\ref{fig:appendix-interaction-cases} and~\ref{fig:appendix-interaction-case-terminal} show two representative OSWorld interaction traces. The first case illustrates a LibreOffice Calc workflow in which the agent consults different spreadsheet skills at different stages of table construction. The second case illustrates a terminal file-organization task where branch guidance helps move past an initially brittle command and then verifies the final archive structure.

\section{Broader Impact}
\label{app:broader-impact}

MMSkills are intended to make visual agents more reliable by externalizing reusable multimodal procedural knowledge. Potential benefits include improved desktop automation, reduced repeated trial-and-error interactions, better support for smaller models, and more reusable agent knowledge across GUI and game-like visual environments. At the same time, more capable visual agents may also increase the risk of unwanted automation, misuse in interactive software, or accidental actions in sensitive environments. Multimodal skill packages can also contain screenshots or cropped visual evidence, so their construction should avoid private or proprietary user data unless appropriate consent, filtering, and access controls are in place. In this work, we construct skills from public non-evaluation trajectories and store compact state evidence rather than raw demonstrations whenever possible. Future deployments should combine MMSkills with permission controls, task-level safety policies, sensitive-information filtering, and auditing of generated skill packages before they are made available to autonomous agents.

\section{Use of LLMs}
\label{app:use-of-llms}

Large language models are used in this work as both research artifacts and research assistants. Methodologically, LLM-based agents are used in the skill-generation pipeline to process and filter trajectories, propose reusable procedures, draft state cards, and generate multimodal skill packages under human-designed schemas and quality checks. LLMs also serve as the evaluated visual agents in the benchmark results. In addition, LLM tools were used during manuscript preparation for editing, polishing, and organizing written content. The authors remained responsible for experimental design, result interpretation, citation checking, and final paper content.

\clearpage

\section{Detailed Related Work}
\label{app:related-work}

This section provides the expanded related-work discussion summarized in Section~\ref{sec:related-work}.

\paragraph{Skills for agents.}
Skill reuse has a long history in temporal abstraction for reinforcement learning and motor primitives for robotics \citep{sutton1999options,ijspeert2013dmp}. Recent LLM agents have made skills a practical interface for storing and composing procedural knowledge in language-conditioned environments. Early systems connected language models to action by grounding language in affordances \citep{ichter2022saycan}, emitting executable programs \citep{liang2023codepolicies}, or interleaving reasoning and acting \citep{yao2023react}; adjacent code- and tool-agent work studies robust tool-call data loops, search-based code refinement, and adversarial test-case generation \citep{zhang2025looptoolclosingdatatrainingloop,li2025rethinkmctsrefiningerroneousthoughts,li2026atgen}. Reflection mechanisms then made agent behavior more persistent across attempts \citep{shinn2023reflexion}. In open-ended environments, systems such as DEPS, Voyager, and JARVIS-1 showed that large models can use language, stored experience, and self-generated programs to acquire or reuse behaviors over extended task horizons \citep{wang2024deps,wang2023voyager,wang2023jarvis}. These works motivate our focus on procedural reuse, but their reusable knowledge is primarily textual, symbolic, or programmatic.

More recent work treats skills as an explicit substrate for agent improvement. SkillWeaver distills web exploration into reusable API-like skills \citep{zheng2025skillweaver}; CUA-Skill builds a parameterized skill base with execution and composition graphs for computer-using agents \citep{chen2026cuaskill}; SkillX automatically constructs hierarchical skill knowledge bases from agent experience \citep{wang2026skillx}; EvoSkill studies automated skill discovery through failure analysis in multi-agent settings \citep{alzubi2026evoskill}, where decentralized coordination and scalable improvement are also central concerns \citep{yang2025agentnetdecentralizedevolutionarycoordination,shao2026monoscalescalingmultiagentmonotonic}; SkillClaw evolves shared skills from multi-user trajectories \citep{ma2026skillclawletskillsevolve}; and SkillRL co-evolves a hierarchical skill library with reinforcement learning \citep{xia2026skillrlevolvingagentsrecursive}. A recent survey frames agent skills as portable packages of instructions, code, and resources loaded through progressive disclosure \citep{xu2026agentskills}. A complementary perspective treats accumulated agent experience as long-term memory: Generative Agents maintain a memory stream that supports recall, reflection, and planning \citep{park2023generativeagentsinteractivesimulacra}, while MemGPT introduces an OS-style memory hierarchy that pages information in and out of the model's working context \citep{packer2024memgptllmsoperatingsystems}. MMSkills follows this broader move toward modular procedural knowledge, but changes the unit being stored: instead of treating skills mainly as text, code, APIs, or execution graphs, we define a skill package whose central evidence is a set of visually grounded runtime states. Branch loading also takes inspiration from memory-paging ideas, by inspecting selected multimodal evidence in a temporary branch rather than flooding the main context.

This emerging ecosystem has also motivated dedicated evaluation of skill utility. SkillsBench measures how skills affect agent performance across diverse tasks \citep{li2026skillsbenchbenchmarkingagentskills}, SkillTester evaluates utility and security risks of agent skills \citep{wang2026skilltesterbenchmarkingutilitysecurity}, and recent work studies skill usage under more realistic retrieval and adaptation settings \citep{liu2026agenticskillsworkwild}. These benchmarks show that skills are not automatically beneficial; their value depends on relevance, compactness, selection, and safe use, especially as self-evolving agents may introduce emergent risks \citep{shao2026agentmisevolveemergentrisks}. Our work addresses a complementary question for visual agents: what evidence should a skill expose, and how should that evidence be loaded, when correct use depends on the current visual state?

The closest line to our work is multimodal and GUI-specific skill augmentation. Mirage-1 introduces hierarchical multimodal skills for GUI agents and uses them with search to support long-horizon control \citep{xie2025mirage}; XSkill continually extracts experiences and skills for multimodal agents from visually grounded rollouts \citep{jiang2026xskillcontinuallearningexperience}; MuSEAgent studies stateful experiences for multimodal reasoning agents \citep{wang2026museagentmultimodalreasoningagent}; and CUA-Skill builds computer-use skills as parameterized procedures and execution graphs \citep{chen2026cuaskill}. MMSkills differs in emphasis: we define the skill artifact around reusable visual state evidence, not only around executable procedure structure or memory accumulation. Each skill is organized around when-to-use conditions, visible cues, verification cues, and multi-view state evidence, and the runtime first selects the relevant evidence before exposing it to the main agent. This makes the contribution a representation and loading mechanism for multimodal procedural cues, rather than another text skill library or GUI action graph.

\paragraph{Visual agents.}
Visual agents have rapidly advanced from web navigation to general computer use. Benchmarks such as Mind2Web and WebArena established realistic web-agent evaluation beyond synthetic interfaces \citep{deng2023mind2web,zhou2024webarena}; VisualWebArena showed that many web tasks require visual grounding rather than text-only reasoning \citep{koh2024visualwebarena}; and WebVoyager demonstrated end-to-end web interaction with large multimodal models on real websites \citep{he2024webvoyager}. The same trend appears in mobile, desktop, and embodied settings: Android in the Wild and AndroidWorld study device control from visual UI observations \citep{rawles2023androidwild,rawles2025androidworld}, OSWorld and macOSWorld evaluate agents in real operating-system environments \citep{xie2024osworld,yang2025macosworld}, RiOSWorld evaluates risks in multimodal computer-use agents \citep{yang2025riosworldbenchmarkingriskmultimodal}, and VisualAgentBench includes VAB-Minecraft and VAB-OmniGibson for open-world and household embodied interaction \citep{liu2024visualagentbench}.

Model and framework work has likewise moved toward visually grounded action, reflecting the shared multimodal objective of aligning visual and textual representations \citep{liu2024alignrecaligningtrainingmultimodalrecommendations,zhang2024dreamdualrepresentationlearningmodel}. SeeClick trains GUI grounding for screenshot-only agents \citep{cheng2024seeclick}; CogAgent introduces a visual language model dedicated to GUI understanding and operation \citep{hong2024cogagentvisuallanguagemodel}; OS-ATLAS learns a foundation action model for GUI control \citep{wu2024osatlas}; UI-TARS develops native GUI agents that perceive screenshots and emit keyboard/mouse actions \citep{qin2025uitars}; SeeAct builds web agents around general-purpose vision-language models \citep{zheng2024gpt4visiongeneralistwebagent}; AppAgent learns smartphone skills from on-device demonstrations \citep{zhang2023appagentmultimodalagentssmartphone}; OmniParser provides a pure-vision parser that turns screenshots into structured GUI elements \citep{lu2024omniparserpurevisionbased}; and Agent S provides a general computer-use framework built around GUI interaction \citep{agashe2024agents}. These systems improve the agent's perceptual and action interface. MMSkills instead targets the external knowledge layer used by such agents. A stronger GUI action model may click more accurately, but it still benefits from knowing which procedural state matters, which visual cue confirms progress, and which state indicates that a skill should not be applied. MMSkills represents that knowledge as a compact, reusable multimodal skill package.

\paragraph{GUI grounding benchmarks.}
Alongside task-completion benchmarks, a separate line of work measures how reliably GUI agents can localize UI elements from natural-language instructions. ScreenSpot-Pro extends earlier ScreenSpot evaluations to high-resolution, professional desktop environments, where target elements often occupy less than $0.1\%$ of the screen and the strongest grounding models still fall well below human performance \citep{li2025screenspotproguigroundingprofessional}. \citet{gou2025navigatingdigitalworldhumans} push toward universal visual grounding that lets agents identify GUI elements purely from screenshots, in the spirit of how humans navigate digital interfaces. MMBench-GUI organizes evaluation hierarchically, from content understanding and element grounding to task automation and multi-agent collaboration \citep{wang2025mmbenchguihierarchicalmultiplatformevaluation}, and DeskVision contributes a large-scale desktop dataset and evaluation suite that broadens grounding research across operating systems \citep{xu2025deskvisionlargescaledesktop}. These benchmarks isolate the perceptual layer of visual agents. MMSkills is complementary: rather than improving where to click, it provides procedural and visual evidence about which state matters at each step, and lets the underlying grounding capability translate that evidence into precise actions.

\paragraph{Long-context reliability.}
Recent studies have shown that simply enlarging the context window does not guarantee that all evidence is used effectively. \citet{liu2023lostmiddlelanguagemodels} report that language models often fail to retrieve information placed in the middle of long contexts, and benchmarks such as LongBench reveal substantial degradation as the input grows in length and modality \citep{bai2024longbenchbilingualmultitaskbenchmark}. These observations motivate our branch-loaded design: rather than directly inserting state cards, multi-view keyframes, and transition examples into the main agent context, the runtime first inspects selected evidence in a temporary branch and returns a compact structured guidance tuple. This isolates expensive multimodal evidence reading from action generation, and avoids the long-context failure modes that arise when reference views and live observations compete for the same context window.

\end{document}